\pdfoutput=1
\PassOptionsToPackage{table}{xcolor}
\documentclass[10pt,twocolumn,letterpaper]{article}

\usepackage[pagenumbers]{cvpr}

\usepackage{currfile}%

\usepackage{caption}%

\usepackage{booktabs}
\usepackage{graphicx}
\usepackage{amsmath,amssymb}
\usepackage{xcolor}
\usepackage{xspace}
\usepackage{multirow}
\usepackage{microtype}
\newcommand{\bench}{ReelCodeBench\xspace}
\newcommand{\Srecon}{S_{\mathrm{recon}}}
\newcommand{\Sedit}{S_{\mathrm{edit}}}
\newcommand{\where}{\textup{\textsc{Where}}\xspace}
\newcommand{\when}{\textup{\textsc{When}}\xspace}
\newcommand{\how}{\textup{\textsc{How}}\xspace}
\usepackage{pifont}
\usepackage{makecell}

\makeatletter
\@ifpackageloaded{lineno}{%
  \IfFormatAtLeastTF{2025-06-01}{%
    \IfPackageAtLeastTF{lineno}{2026/02/16}{%
      \IfPackageAtLeastTF{lineno}{2026/02/17}{}{%
        \RemoveFromHook{build/column/before}[lineno]%
        \AddToHook{build/column/after}{%
          \setbox\@outputbox\vbox{%
            \boxmaxdepth\@maxdepth
            \protected@write\@auxout{}{%
              \string\def\string\@LN@column{\if@firstcolumn1\else2\fi}}%
            \box\@outputbox
          }%
        }%
      }%
    }{}%
  }{}%
}{}%
\makeatother
\usepackage{enumitem}
\usepackage[T1,OT1]{fontenc}
\DeclareTextSymbolDefault{\textquotedbl}{T1}
\usepackage{listings}
\usepackage[most]{tcolorbox}

\definecolor{cvprblue}{rgb}{0.21,0.49,0.74}
\usepackage[pagebackref,breaklinks,colorlinks,allcolors=cvprblue]{hyperref}
\usepackage{tabularx}
\usepackage{placeins}

\definecolor{rcol}{HTML}{9B1C1C}
\definecolor{gcol}{HTML}{2E7D4F}
\definecolor{ucol}{HTML}{1F5FA8}
\definecolor{ccol}{HTML}{B7791F}
\definecolor{xcol}{HTML}{4A5568}
\definecolor{phc}{HTML}{9B1C1C}
\newcommand{\phl}[1]{\textcolor{phc}{\textbf{\{#1\}}}}
\newcommand{\rk}[1]{\,{\scriptsize\color{gray}(#1)}}

\tcbset{promptstyle/.style={enhanced,listing only,colback=#1!4,colframe=#1,coltitle=white,fonttitle=\bfseries\small,
  colbacktitle=#1,arc=2mm,boxrule=0.8pt,left=4pt,right=4pt,top=3pt,bottom=3pt,
  listing options={basicstyle=\ttfamily\scriptsize,breaklines=true,columns=fullflexible,keepspaces=true,upquote=true,escapeinside={(*@}{@*)}}}}

\title{ReelCodeBench: Can Frontier LLMs Edit What They Rebuild?}

\author{%
Hohin Kwan$^{1}$\quad
Hongyu Li$^{2,\dagger}$\quad
Hao Yu$^{3}$\quad
Chi Hsu Tsai$^{4}$\\
Pei Yang$^{1}$\quad
Houyuan Chen$^{1}$\quad
Si Liu$^{2}$\quad
Anyi Rao$^{1,*}$\\[0.2em]
{\small $^{1}$HKUST MMLab\quad
$^{2}$Beihang University\quad
$^{3}$Tsinghua University\quad
$^{4}$Peking University}\\[0.1em]
{\small $^{\dagger}$Project leader\quad $^{*}$Corresponding author}\\
{\small Project Page: \url{https://mrakas.github.io/reelcode}}\\[-14pt]
}

\begin{document}
\makeatletter
\AddToHookNext{cmd/@maketitle/after}{\begin{minipage}{\textwidth}
  \centering
  \captionsetup{type=figure}
  \includegraphics[width=\textwidth]{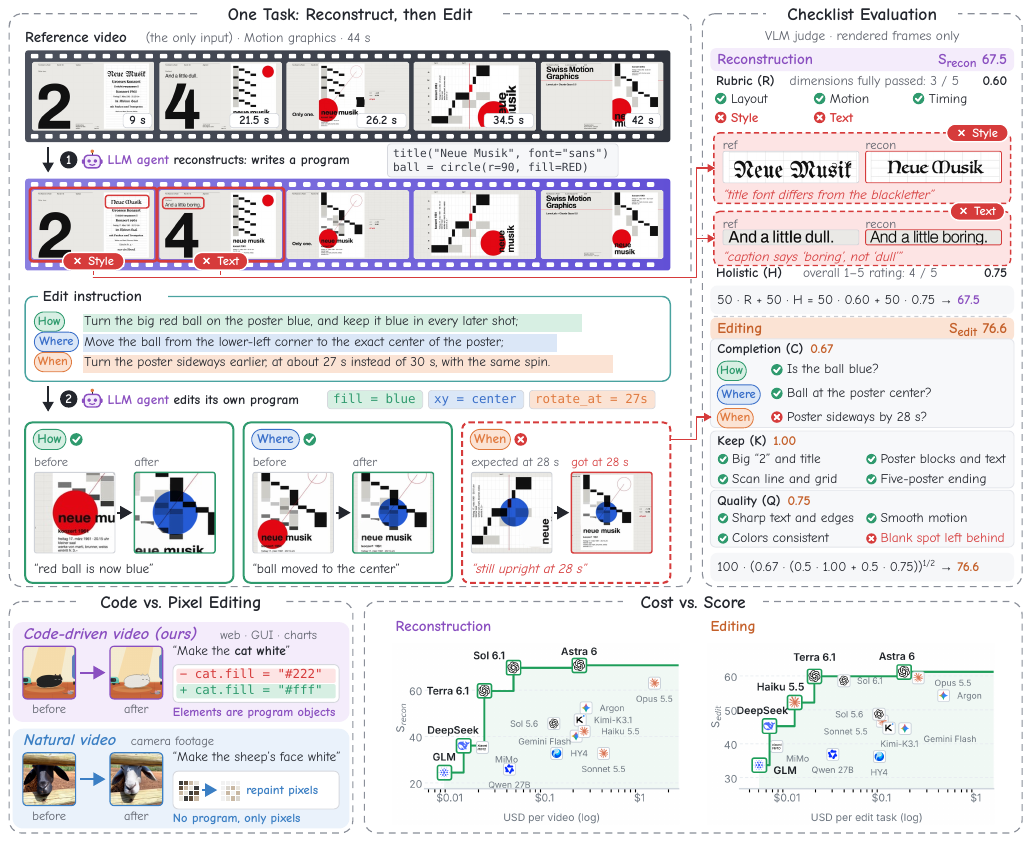}
  \caption{\textbf{Reconstructing and editing video with code.}
  A model recreates a reference video as a program, then edits its own code on request.
  The judge checks rendered frames (right).
  The bottom panels compare code and pixel editing and show the cost--score Pareto fronts.}
  \label{fig:teaser}
\end{minipage}
}
\makeatother
\maketitle
\raggedbottom
\begin{abstract}
Large language models increasingly create visual content by writing code, and motion-graphics videos are a natural target: many are already produced as programs in HTML, SVG, Canvas, or Three.js.
Code makes a video editable, yet no benchmark tests whether a model can recreate a video from pixels alone and then edit its own program on request.
We introduce \bench, a benchmark for code-based video reconstruction and editing.
Given only a reference video, a model writes a program that reproduces it, and then edits that program following a natural-language instruction.
\bench contains 420 videos, ranging from few-second motion clips to three-minute explainer animations across six content types and six visual forms, each paired with an editing task, for 844 edit requirements organized into a \where\,/\,\when\,/\,\how taxonomy of spatial, temporal, and appearance edits.
Since edited videos have no ground truth, we score each task with a rubric answered by a vision-language model from rendered frames: completion questions for each change inside its time window, keep items for content that must stay unchanged, and an overall quality rating.
This separates missed edits, wrong edits, and collateral damage.
Evaluating sixteen code-generating models and two diffusion video generation models, we find that reconstruction errors carry into editing, that motion and timing are the main bottleneck in both stages, and that even the best editor keeps only about 60\% of what it should.
\end{abstract}

\section{Introduction}
\label{sec:intro}

Imagine a content creator who wants to reuse a striking motion-graphics intro found online but has only the video: the practical route is to recreate it as a program and then adjust its colors, layout, and timing.
This recreate-then-edit workflow is common in the production of motion graphics, UI animations, and data visualizations.
Unlike videos synthesized in pixel space by diffusion models~\cite{wan2025}, code videos written in HTML/CSS, SVG, Canvas, or Three.js expose the position, appearance, and timing of every element as explicit program variables, and therefore naturally support precise local edits.
With the rapid progress of multimodal large language models (MLLMs) in code generation, carrying out this workflow with a model is becoming feasible~\cite{chen2025code2video,chen2026compass,chen2026vibeprevisbench,wang2026internsvg,yang2025omnisvg}.

Despite this progress, existing benchmarks still suffer from three major challenges. \textbf{i)~Decoupled reconstruction and editing:} Prior work decouples visual reconstruction from downstream code editing. Video-to-code benchmarks~\cite{chen2026compass,ji2026animation2code,dai2026webvr,zhang2026animweb,diao2026vibevideobench} stop at one-shot visual reproduction, whereas code-editing benchmarks~\cite{zhao2025chartedit,designbench,chen2025svgenius,editbench2025,gu2025blendergym} start from clean, human-written source code. No benchmark asks a model to edit the imperfect program that it has itself reconstructed from raw pixels.~\mbox{\textbf{ii) Code-editing tasks remain confined to static visuals:}} Most visual code-editing benchmarks~\cite{zhao2025chartedit,tang2025chart2code,designbench,gu2025blendergym} edit static charts, web pages, or 3D scenes, so their instructions change appearance or layout but not when events happen. MotionGraphicsBench~\cite{ma2026motiongraphicsbench} adds motion edits, but on supplied SVG code rather than on a program the model rebuilt from video.~\textbf{iii) Fragile evaluation metrics:} Video editing benchmarks~\cite{chen2025ivebench,liang2026goku} measure fidelity by global similarity to the source video, which penalizes requested and unintended changes alike, while the checklists of recent code-video benchmarks~\cite{chen2026compass,diao2026vibevideobench} verify only what should appear. Neither can tell whether an edit ignored the instruction, changed the wrong element or moment, or damaged other content.
Therefore, there is an urgent need for a benchmark that couples reconstruction with self-editing, covers spatial, temporal, and appearance edits, and diagnoses why an edit fails.

\begin{figure*}[!t]
  \centering
  \includegraphics[width=\textwidth]{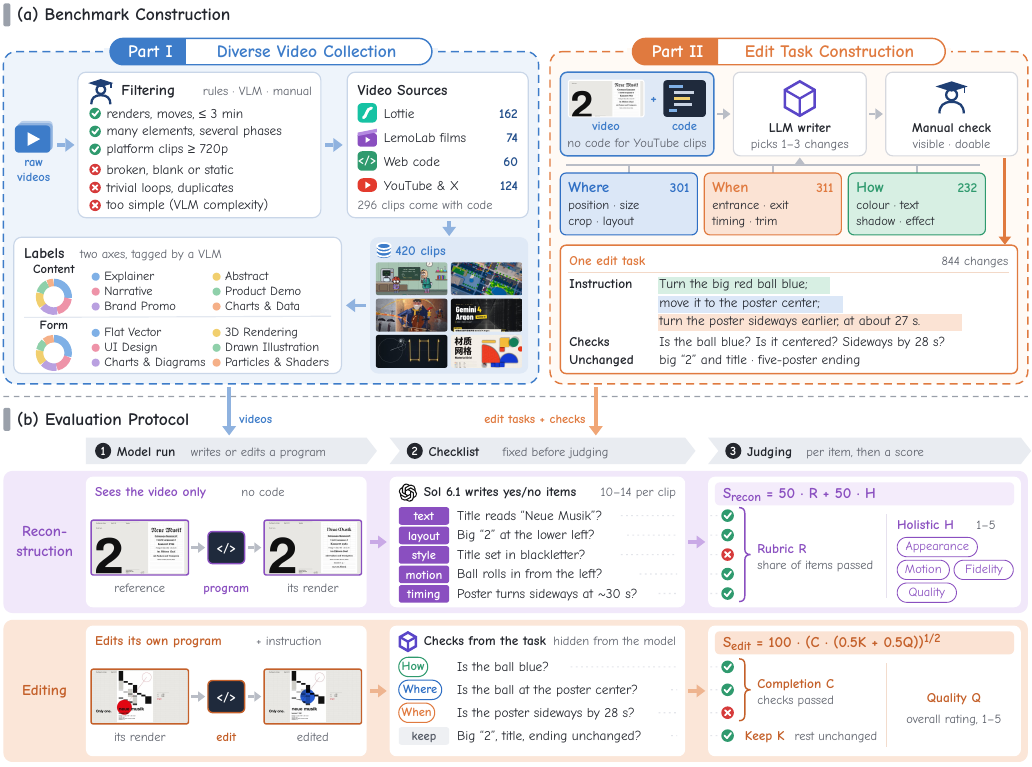}
  \caption{\textbf{Benchmark construction and evaluation.}
  (a) We curate and label motion-graphics clips and manually verify the edit tasks.
  (b) VLM judges score rendered frames using fixed checklists for reconstruction and editing.}
  \label{fig:pipeline}
\end{figure*}

\begin{figure}[!t]
  \centering
  \includegraphics[width=\columnwidth]{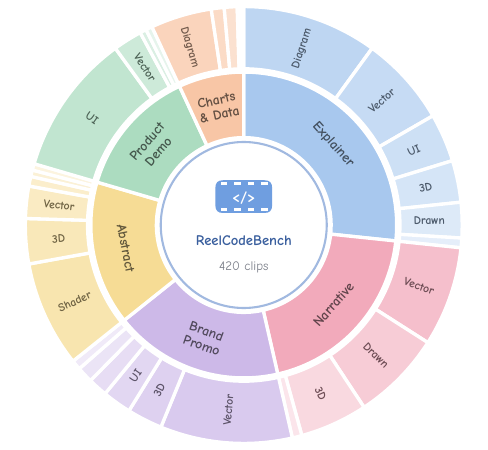}
  \caption{\textbf{Content and visual form of the clips.}
  Inner ring: content type; outer ring: visual form.
  More statistics are in \cref{sec:s_data}.}
  \label{fig:stats}
\end{figure}

To this end, we propose \textbf{\bench}, a benchmark for reconstructing and editing videos with code, named after the short \emph{reels} that make up most of its data.
Addressing the three challenges above, \bench features three key designs.
\mbox{\textbf{1) Coupled reconstruction and self-editing:}}
Given only a reference video, a model first writes a program that recreates it, and then modifies its own program following a natural-language instruction.
The benchmark contains 420 motion-graphics videos from 5 sources, spanning 6 content types and 6 visual forms, with durations from 3 seconds to 3 minutes (\cref{fig:teaser,fig:stats}).
\mbox{\textbf{2) A unified \where/\when/\how edit taxonomy:}}
Each video comes with one human-verified editing task whose requirements span spatial layout (\where), timing (\when), and visual appearance (\how), including timing edits that static code-editing benchmarks do not cover.
\mbox{\textbf{3) A decomposed diagnostic evaluation:}}
Reconstruction is scored with per-video binary rubrics over five dimensions (text, layout, style, motion, and timing) plus a holistic quality score.
Editing is decomposed into completion $C$, keep rate $K$, and quality $Q$: completion is checked within the time window of each requested change, and keep rate checks the content that should remain unchanged (\cref{fig:pipeline}).
Because each requested change and each preserved element is checked separately, the three failure modes above are distinguished. All scores are averaged over two judges from different model families, and none requires a target edited video.

The main contributions of this work are:
\begin{itemize}[leftmargin=*,nosep]
\item \textbf{Novel Task and Benchmark:} We introduce \bench, a benchmark for video-to-code reconstruction followed by self-editing, with edits organized under a unified \where/\when/\how taxonomy.
\item \textbf{Diagnostic Evaluation Protocol:} We score reconstruction with rubrics plus holistic ratings, and editing by completion, keep rate, and quality, so missed edits and collateral damage are scored apart.
\item \textbf{Extensive Benchmarking and Insights:} We evaluate 16 code-generating models and diffusion video generation baselines, revealing bottlenecks in motion, code execution, and keeping unedited content intact.
\end{itemize}
We will release the data and evaluation code.

\newcommand{\mlogo}[1]{\raisebox{-0.18em}{\includegraphics[height=1.05em]{fig/logo_#1.pdf}}\hspace{0.3em}}
\newcommand{\LogoClaude}{\mlogo{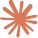}}\newcommand{\LogoOpenAI}{\mlogo{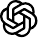}}\newcommand{\LogoKimi}{\mlogo{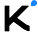}}
\newcommand{\LogoDeepSeek}{\mlogo{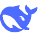}}\newcommand{\LogoMiMo}{\mlogo{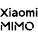}}\newcommand{\LogoQwen}{\mlogo{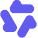}}
\newcommand{\LogoGLM}{\mlogo{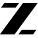}}\newcommand{\LogoSenseNova}{\mlogo{sensenova}}\newcommand{\LogoMiniMax}{\mlogo{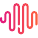}}\newcommand{\LogoGemini}{\mlogo{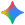}}\newcommand{\LogoHunyuan}{\mlogo{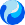}}\newcommand{\LogoWan}{\mlogo{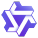}}
\definecolor{rk1}{RGB}{82,214,82}
\definecolor{rk2}{RGB}{160,240,140}
\definecolor{rk3}{RGB}{218,250,210}
\definecolor{bst}{gray}{0.66}
\definecolor{snd}{gray}{0.89}
\definecolor{grp}{RGB}{238,244,250}
\definecolor{bandA}{RGB}{253,236,224}
\definecolor{bandB}{RGB}{255,249,222}
\definecolor{bandC}{RGB}{228,242,232}
\newcommand{\rh}[1]{\textit{#1}}

\section{\bench}
\label{sec:bench}

\bench evaluates whether a model can reconstruct a reference video with code and then edit it following an instruction.
We define the two stages (\cref{sec:task}), then describe how we collect videos (\cref{sec:data}) and build the editing tasks (\cref{sec:construct}), and summarize the data (\cref{sec:stats}).

\subsection{Task Definition}
\label{sec:task}
\label{sec:task-recon}
\label{sec:task-edit}

\paragraph{Video Reconstruction.}
Given a reference video $V$, a model $M$ writes an executable program $P=M(V)$, which a renderer $R$ records into the reconstruction $\hat V=R(P)$.
The model sees only the video, never its source code.
$\hat V$ should match $V$ at every moment in what is shown, where it is, how it looks, and how and when it moves~\cite{hong2025motionbench} (\cref{sec:recon-score}).
The model is free to choose how to render, for example with CSS, SVG, Canvas, or Three.js; crashing or blank programs fail.

\paragraph{Video Editing.}
Given a natural-language editing instruction $I$, the model modifies its own program, $P'=M(V,\hat V,P,I)$, which renders to the edited video $\hat V'=R(P')$.
Each instruction contains one to three edit requirements.
A good edit completes every requirement correctly, leaves content not mentioned in the instruction unchanged, and keeps the edited video intact (\cref{sec:edit-score}).
Because the model edits its own reconstruction, an element that is missing or poorly structured in $P$ is also hard to edit.

\subsection{Video Collection}
\label{sec:data}

\paragraph{Sources.}
We start from a pool of about 5.8K candidate motion-graphics videos and keep those that pass filtering, drawn from five sources (\cref{fig:pipeline}a, Part I):
162 Lottie animations from LottieFiles,
74 code-driven short films from LemoLab~\cite{lemolab},
60 web animations from open-source repositories,
87 clips from YouTube and other video sites, and
37 clips from X.
YouTube and X clips are video only; all other sources provide the original code, which we render.

\paragraph{Filtering.}
We filter candidates first by rules and then with a VLM.
(1)~\emph{Rule-based filtering.} We keep videos between about 3\,s and 3\,min and drop those that are low in resolution, static, blank, or short trivial loops. For videos with code, we further require at least 30 lines of code excluding third-party libraries and at least three independently moving regions. Near-duplicates are removed with perceptual hashes of key frames, and each author contributes at most five videos.
(2)~\emph{VLM filtering.} A VLM (Sol 6.1~\cite{gpt61}) first labels each remaining video with one content type and one visual form, and we sample across these labels to keep the set diverse. It then inspects frames sampled at 1\,fps, plus code when available, and drops broken or trivial videos.

\subsection{Task Construction}
\label{sec:construct}

\paragraph{Reconstruction tasks.}
Each reconstruction task is a reference video that the model reproduces as an executable program.
For videos with source code, we render the reference from its original code; for video-only clips, we use the collected clip directly.
In both cases the model receives only the reference video, never its source code.

\paragraph{Editing tasks.}
Each editing task is a natural-language instruction that combines one to three edit requirements.
For videos with source code, an LLM (Sol 6.1) writes the instruction from both the video and its code; for video-only clips, it writes the instruction from the video alone.
Every task is then reviewed by human experts.

\paragraph{Edit taxonomy.}
An edit requirement modifies one visible element within one time window, and falls into one of three types:
\textbf{\where} changes spatial properties such as position, size, crop, and layout;
\textbf{\when} changes behavior over time, such as appearance time, entrance and exit, motion, transitions, and trimming; and
\textbf{\how} changes appearance, such as text, typography, color, and effects.

\subsection{Statistics}
\label{sec:stats}

\looseness=-1 \Cref{fig:stats} shows the content and visual form of the clips; \cref{sec:s_data} gives the other statistics.
Each video comes with one editing instruction, giving 844 edit requirements in total.
The videos span six content types (Explainer, Narrative, Brand Promo, Abstract, Product Demo, Charts \& Data) and six visual forms (Flat Vector, UI Design, Charts \& Diagrams, 3D Rendering, Drawn Illustration, Particles \& Shaders).
Durations range from 2.9\,s to 179.3\,s, with a median of 20\,s.
An instruction contains one, two, or three requirements (123 / 170 / 127 instructions), and the three types are roughly balanced: 301 \where, 311 \when, and 232 \how.
On average, each video is evaluated with 21.8 binary criteria: 13.9 reconstruction rubric items, 4.1 editing checks, and 3.8 keep items.

\begin{table*}[!t]\centering
\caption{\textbf{Reconstruction results.}
$S_\mathrm{recon}$ combines the rubric score $R$ and the holistic score $H$, the mean of the four ratings (\cref{sec:recon-score}).
Objective columns report frame similarity ($\times 100$).}\label{tab:recon}
\small\setlength{\tabcolsep}{2.9pt}\renewcommand{\arraystretch}{1.12}
\resizebox{\textwidth}{!}{%
\begin{tabular}{l|cc|cccccccccccccc}
 & & & \multicolumn{6}{c}{\cellcolor{bandA}Rubric (dimension fully passed)} & \multicolumn{5}{c}{\cellcolor{bandB}Holistic} & \multicolumn{3}{c}{\cellcolor{bandC}Objective} \\
Methods & Rank & $S_\mathrm{recon}$ & \rh{$R$} & \rh{Text} & \rh{Layout} & \rh{Style} & \rh{Motion} & \rh{Timing} & \rh{$H$} & \rh{Appear.} & \rh{Motion} & \rh{Fidelity} & \rh{Quality} & \rh{SSIM} & \rh{DINOv2} & \rh{CLIP} \\ \hline
\rowcolor{grp}\multicolumn{17}{l}{\textit{Closed-source Models (API)}} \\
\LogoOpenAI{}GPT-6 Astra & \cellcolor{rk1}1 & 70.7 & \cellcolor{bst}72.2 & \cellcolor{bst}79.0 & 80.5 & \cellcolor{bst}79.5 & 65.2 & \cellcolor{bst}73.4 & \cellcolor{bst}69.2 & \cellcolor{bst}68.8 & \cellcolor{bst}66.1 & \cellcolor{bst}66.3 & \cellcolor{bst}75.6 & \cellcolor{bst}79.0 & \cellcolor{bst}80.1 & \cellcolor{bst}88.1 \\
\LogoOpenAI{}GPT-6.1 Sol & \cellcolor{rk2}2 & 69.7 & 71.2 & 77.0 & \cellcolor{bst}82.7 & 75.6 & \cellcolor{bst}65.5 & 71.3 & 68.2 & 67.5 & 65.5 & 65.5 & 74.5 & 78.3 & 80.0 & 87.3 \\
\LogoClaude{}Claude Opus 5.5 & \cellcolor{rk3}3 & 63.0 & 61.5 & 75.4 & 75.4 & 74.5 & 49.5 & 58.0 & 64.5 & 64.2 & 58.9 & 60.4 & 74.5 & 75.8 & 78.5 & 87.9 \\
\LogoOpenAI{}GPT-6.1 Terra & 4 & 59.7 & 57.4 & 59.1 & 67.9 & 70.7 & 49.6 & 60.2 & 61.9 & 60.6 & 61.1 & 59.3 & 66.6 & 77.0 & 77.7 & 86.6 \\
\LogoGemini{}Gemini 4 Argon & 5 & 52.3 & 49.2 & 53.7 & 60.8 & 65.9 & 41.3 & 47.0 & 55.5 & 56.2 & 52.2 & 52.3 & 61.2 & 71.7 & 73.5 & 81.8 \\
\LogoOpenAI{}GPT-5.6 Sol & 7 & 45.5 & 39.3 & 57.8 & 45.7 & 54.3 & 27.3 & 42.3 & 51.7 & 50.1 & 46.2 & 46.6 & 63.7 & 71.3 & 72.1 & 83.9 \\
\LogoClaude{}Claude Haiku 5.5 & 8 & 42.2 & 34.2 & 54.9 & 48.6 & 48.9 & 22.6 & 32.7 & 50.2 & 50.8 & 41.2 & 43.8 & 64.9 & 75.2 & 74.5 & 86.8 \\
\LogoGemini{}Gemini 3.8 Flash & 9 & 40.1 & 32.3 & 47.9 & 42.0 & 49.1 & 21.5 & 33.7 & 47.9 & 46.8 & 41.5 & 42.5 & 60.8 & 67.6 & 70.2 & 81.4 \\
\LogoOpenAI{}GPT-6 Luna & 10 & 39.8 & 33.8 & 49.7 & 40.3 & 50.5 & 22.7 & 37.0 & 45.8 & 44.6 & 42.0 & 41.3 & 55.3 & 70.3 & 71.0 & 81.8 \\
\LogoClaude{}Claude Sonnet 5.5 & 14 & 32.7 & 23.9 & 39.3 & 39.9 & 47.2 & 10.6 & 19.0 & 41.5 & 44.0 & 27.5 & 32.8 & 61.6 & 74.5 & 72.0 & 85.0 \\
\hline
\rowcolor{grp}\multicolumn{17}{l}{\textit{Open-weight Models}} \\
\LogoKimi{}Kimi-K3.1-Pro & 6 & 47.1 & \cellcolor{snd}42.4 & \cellcolor{snd}47.3 & \cellcolor{snd}56.0 & \cellcolor{snd}51.2 & \cellcolor{snd}36.3 & \cellcolor{snd}42.6 & \cellcolor{snd}51.7 & \cellcolor{snd}50.4 & \cellcolor{snd}50.1 & \cellcolor{snd}48.5 & 57.7 & \cellcolor{snd}69.9 & \cellcolor{snd}70.4 & 79.6 \\
\LogoDeepSeek{}DeepSeek-V4.1-Flash (1017) & 11 & 36.0 & 26.8 & 44.7 & 34.5 & 40.1 & 15.4 & 33.8 & 45.2 & 42.8 & 35.8 & 37.7 & \cellcolor{snd}64.4 & 63.2 & 66.9 & \cellcolor{snd}79.8 \\
\LogoMiMo{}MiMo-V2.6-Flash & 12 & 35.3 & 28.2 & 40.6 & 39.2 & 44.5 & 21.0 & 26.3 & 42.5 & 41.5 & 36.9 & 36.7 & 54.9 & 60.1 & 63.5 & 74.2 \\
\LogoHunyuan{}HY4-preview & 13 & 32.7 & 29.1 & 32.5 & 36.1 & 34.5 & 26.6 & 32.9 & 36.3 & 34.4 & 36.6 & 34.0 & 40.3 & 49.1 & 48.9 & 56.2 \\
\LogoQwen{}Qwen3.8-27B & 15 & 25.9 & 16.9 & 31.2 & 24.6 & 29.1 & 10.5 & 18.5 & 35.0 & 33.7 & 27.7 & 28.8 & 49.6 & 61.7 & 61.0 & 73.4 \\
\LogoGLM{}GLM-5.3-Flash & 16 & 24.4 & 15.5 & 28.4 & 20.6 & 34.0 & 9.8 & 14.2 & 33.2 & 32.9 & 26.0 & 26.0 & 48.1 & 62.4 & 62.3 & 74.8 \\
\hline
\end{tabular}}
\end{table*}

\section{Evaluation Protocol}
\label{sec:eval}

We score both stages with rubrics that a vision-language model (VLM) judge answers from rendered frames (\cref{fig:pipeline}b), in the spirit of question-based checks for visual generation and editing~\cite{hu2023tifa,cho2024dsg,li2025fivebench}.

\subsection{Reconstruction Score}
\label{sec:recon-score}

\paragraph{Evaluation dimensions.}
We score a reconstruction from two sides.
(1)~Rubric items cover five dimensions:
\textbf{text}, the words on screen;
\textbf{layout}, the position, size, and stacking of elements;
\textbf{style}, colors, fonts, materials, and visual effects;
\textbf{motion}, how elements move, scale, rotate, and deform; and
\textbf{timing}, when elements appear and disappear and when each animation segment happens.
(2)~Holistic ratings score the whole reconstruction from 1 to 5 on appearance, motion, fidelity, and visual quality.

\paragraph{Rubric generation.}
For each video, Sol 6.1 generates 10--14 binary criteria from its frames, each checking one visible fact in one of the five dimensions and weighted by importance: 0.8--1.0 for facts without which a viewer would call it a different video, 0.4--0.7 for important facts, and 0.1--0.3 for minor ones. At least three criteria cover motion or timing.
Holistic ratings use four fixed, shared questions.

\paragraph{Scoring.}
The judge compares reference and reconstruction frames sampled at 1\,fps at matching timestamps.
An item passes only on a full match, and a dimension passes only if all of its items pass.
The rubric score $R$ is the share of total item weight that lies in passed dimensions, and the holistic score $H$ is the mean of the four ratings, normalized to 0--1:
\begin{equation}
  \Srecon = 50\cdot R + 50\cdot H.
  \label{eq:recon}
\end{equation}

\subsection{Editing Score}
\label{sec:edit-score}

\paragraph{Evaluation dimensions.}
Following image-editing benchmarks~\cite{ku2024viescore,ye2025imgedit,zhao2025risebench}, we score an edit on three separate dimensions:
(1)~\textbf{completion} $C$, how fully the requirements are done;
(2)~\textbf{keep rate} $K$, unmentioned content stays unchanged; and
(3)~\textbf{quality} $Q$, the edited video stays clean.

\paragraph{Rubric generation.}
Together with each editing instruction, Sol 6.1 generates its rubric: one to three binary checks per requirement, answerable from the edited video, for $C$, and two to four keep items, nearby elements that are easy to change by mistake, for $K$. Quality uses one shared question.

\paragraph{Scoring.}
The judge compares frames of the reference video, sampled at 1\,fps, with the matching frames of the edited render, so keep items are checked against the reference itself.
$C$ is the mean completion over requirements (0 if nothing visibly changed), $K$ the share of keep items kept, and $Q$ the quality rating, all normalized to 0--1.
As in VIEScore~\cite{ku2024viescore}, we take the geometric mean of completion and the average of keep rate and quality, so an edit that changes nothing scores zero however well it keeps the rest:
\begin{equation}
  \Sedit = 100\cdot\sqrt{C\cdot\left(0.5\,K + 0.5\,Q\right)}.
  \label{eq:edit}
\end{equation}

\subsection{Judges and Aggregation}

We use two VLM judges from different model families, Sol 6.1~\cite{gpt61} and Sonnet 5.5~\cite{sonnet55}, with the same prompts, and report the mean of their scores, following panel-of-judges evaluation~\cite{verga2024juries}.
Failures of any kind, including programs that do not run, blank renders, and truncated outputs, score 0, so every model is averaged over the full benchmark.
For reconstruction we also report three objective similarities between reference and reconstruction frames at matching timestamps: SSIM~\cite{wang2004ssim} on grayscale frames, and the cosine similarity of DINOv2~\cite{oquab2023dinov2} and CLIP~\cite{radford2021clip} image embeddings, each averaged over frame pairs sampled at 1\,fps.
\Cref{sec:s_results,sec:s_rel} give detailed and per-judge results.

\begin{table*}[!t]
\centering
\definecolor{bandAll}{RGB}{232,235,241}
\definecolor{bandWh}{RGB}{220,232,250}
\definecolor{bandWn}{RGB}{253,230,212}
\definecolor{bandHw}{RGB}{218,241,228}

\caption{\textbf{Editing results.}
$S_\mathrm{edit}$ combines completion ($C$), keep rate ($K$), and quality ($Q$).
Under \where, \when, and \how, $C$ is completion of that change type; $K$ and $Q$ average over tasks that contain it.
$^\ddagger$Diffusion video generation models.}
\label{tab:edit}

\small
\setlength{\tabcolsep}{2pt}
\renewcommand{\arraystretch}{1.12}

\begin{tabularx}{\textwidth}{
  @{}l|cc|
  *{3}{*{3}{>{\centering\arraybackslash}X}|}
  *{3}{>{\centering\arraybackslash}X}@{}
}
 & & &
 \multicolumn{3}{c|}{\cellcolor{bandAll}All changes} &
 \multicolumn{3}{c|}{\cellcolor{bandWh}\where} &
 \multicolumn{3}{c|}{\cellcolor{bandWn}\when} &
 \multicolumn{3}{c}{\cellcolor{bandHw}\how} \\
Methods & Rank & $S_\mathrm{edit}$ &
 \rh{$C$} & \rh{$K$} & \rh{$Q$} &
 \rh{$C$} & \rh{$K$} & \rh{$Q$} &
 \rh{$C$} & \rh{$K$} & \rh{$Q$} &
 \rh{$C$} & \rh{$K$} & \rh{$Q$} \\ \hline

\rowcolor{grp}\multicolumn{15}{l}{\textit{Closed-source Models (API)}} \\
\LogoOpenAI{}GPT-6 Astra & \cellcolor{rk1}1 & 61.2
 & \cellcolor{bst}72.1 & \cellcolor{bst}60.8 & 52.8
 & 75.0 & 60.7 & 50.5
 & \cellcolor{bst}63.1 & \cellcolor{bst}64.0 & 52.5
 & \cellcolor{bst}77.0 & \cellcolor{bst}59.2 & 51.6 \\
\LogoOpenAI{}GPT-6.1 Terra & \cellcolor{rk2}2 & 59.9
 & 70.6 & 59.6 & 53.7
 & \cellcolor{bst}77.4 & \cellcolor{bst}61.8 & 52.8
 & 61.7 & 62.1 & 53.0
 & 74.0 & 59.2 & \cellcolor{bst}54.1 \\
\LogoClaude{}Claude Opus 5.5 & \cellcolor{rk3}3 & 59.5
 & 70.5 & 58.1 & \cellcolor{bst}54.0
 & 75.6 & 60.1 & \cellcolor{bst}53.6
 & 61.9 & 60.5 & \cellcolor{bst}53.6
 & 74.7 & 56.4 & 53.1 \\
\LogoOpenAI{}GPT-6.1 Sol & 4 & 58.6
 & 68.2 & 60.1 & 51.2
 & 70.5 & 60.4 & 48.7
 & 60.2 & 62.8 & 51.2
 & 72.1 & 56.8 & 49.4 \\
\LogoGemini{}Gemini 4 Argon & 5 & 54.2
 & 62.5 & 59.6 & 52.9
 & 67.2 & 60.8 & 51.5
 & 54.9 & 61.4 & 52.6
 & 66.5 & 58.4 & 52.1 \\
\LogoClaude{}Claude Haiku 5.5 & 6 & 52.2
 & 63.7 & 52.8 & 49.8
 & 69.6 & 54.2 & 48.9
 & 57.6 & 55.9 & 48.6
 & 64.9 & 51.5 & 49.9 \\
\LogoOpenAI{}GPT-5.6 Sol & 7 & 48.7
 & 57.9 & 53.9 & 47.3
 & 60.1 & 55.9 & 46.4
 & 49.4 & 55.5 & 45.5
 & 63.8 & 53.4 & 48.1 \\
\LogoClaude{}Claude Sonnet 5.5 & 8 & 46.7
 & 57.7 & 52.8 & 46.2
 & 58.4 & 56.3 & 45.5
 & 49.1 & 54.6 & 45.8
 & 66.1 & 50.2 & 46.3 \\
\LogoGemini{}Gemini 3.8 Flash & 11 & 44.5
 & 52.4 & 50.5 & 47.2
 & 54.5 & 50.7 & 46.4
 & 44.7 & 52.5 & 45.7
 & 59.1 & 51.2 & 47.2 \\
\LogoOpenAI{}GPT-6 Luna & 12 & 43.5
 & 52.1 & 52.9 & 46.1
 & 53.5 & 53.9 & 44.5
 & 46.5 & 55.3 & 44.3
 & 54.5 & 52.7 & 46.3 \\
\hline

\rowcolor{grp}\multicolumn{15}{l}{\textit{Open-weight Models}} \\
\LogoDeepSeek{}DeepSeek-V4.1-Flash (1017) & 9 & 45.2
 & \cellcolor{snd}54.5 & 47.6 & \cellcolor{snd}47.6
 & \cellcolor{snd}53.4 & 49.6 & \cellcolor{snd}45.8
 & \cellcolor{snd}47.8 & 50.1 & \cellcolor{snd}47.2
 & \cellcolor{snd}61.6 & 46.3 & \cellcolor{snd}46.5 \\
\LogoKimi{}Kimi-K3.1-Pro & 10 & 44.6
 & 51.6 & \cellcolor{snd}53.0 & 46.4
 & 51.6 & \cellcolor{snd}51.5 & 43.5
 & 46.7 & \cellcolor{snd}56.2 & 46.8
 & 55.5 & \cellcolor{snd}51.0 & 45.6 \\
\LogoMiMo{}MiMo-V2.6-Flash & 13 & 39.2
 & 48.0 & 44.2 & 42.5
 & 51.3 & 46.9 & 42.0
 & 42.9 & 45.1 & 41.3
 & 52.0 & 42.3 & 41.4 \\
\LogoQwen{}Qwen3.8-27B & 14 & 36.9
 & 45.3 & 43.1 & 39.6
 & 47.5 & 43.8 & 38.1
 & 37.8 & 43.6 & 37.0
 & 48.5 & 42.3 & 39.8 \\
\LogoHunyuan{}HY4-preview & 15 & 36.2
 & 46.1 & 31.8 & 32.0
 & 47.9 & 33.9 & 31.7
 & 40.7 & 34.4 & 32.1
 & 49.5 & 30.9 & 32.2 \\
\LogoGLM{}GLM-5.3-Flash & 16 & 33.7
 & 42.1 & 43.5 & 36.9
 & 42.7 & 45.2 & 34.7
 & 36.8 & 43.7 & 35.2
 & 48.1 & 44.1 & 38.3 \\
\hline

\rowcolor{grp}\multicolumn{15}{l}{\textit{Diffusion Video Generation Models}} \\
\LogoMiniMax{}$^\ddagger$MiniMax-H3 & - & 39.5
 & 38.5 & 69.7 & 54.2
 & 27.0 & 68.9 & 50.6
 & 32.0 & 73.3 & 54.7
 & 56.4 & 66.7 & 53.9 \\
\LogoWan{}$^\ddagger$Wan2.1-VACE-1.3B & - & 15.6
 & 17.2 & 48.0 & 27.1
 & 11.9 & 48.2 & 26.2
 & 17.0 & 50.2 & 27.3
 & 22.0 & 47.8 & 25.1 \\
\hline
\end{tabularx}
\end{table*}
\definecolor{allg}{RGB}{238,240,244}
\begin{table*}[!t]\centering
\caption{\textbf{Performance by visual form.} Columns: Flat Vector, Drawn Illustration, 3D Rendering, Particles \& Shaders, UI Design, Charts \& Diagrams.
Blue and orange cells are above and below the model's overall score (All, gray); darker means a larger gap (up to $\pm$25).}\label{tab:bycat}
\footnotesize\setlength{\tabcolsep}{3.2pt}\renewcommand{\arraystretch}{0.95}
\resizebox{\textwidth}{!}{%
\begin{tabular}{l|c|cccccc|c|cccccc}
 & \multicolumn{7}{c|}{\cellcolor{bandA}(a) Reconstruction} & \multicolumn{7}{c}{\cellcolor{bandC}(b) Editing} \\
Methods & All & Vector & Drawn & 3D & Shader & UI & Diagram & All & Vector & Drawn & 3D & Shader & UI & Diagram \\ \hline
\LogoOpenAI{}GPT-6 Astra & \cellcolor{allg}71 & \cellcolor[RGB]{197,214,238}76 & \cellcolor[RGB]{245,213,195}64 & \cellcolor[RGB]{241,197,172}61 & \cellcolor[RGB]{237,180,147}58 & \cellcolor[RGB]{214,226,243}74 & \cellcolor[RGB]{183,205,234}78 & \cellcolor{allg}61 & \cellcolor[RGB]{178,202,233}69 & \cellcolor[RGB]{233,163,123}45 & \cellcolor[RGB]{248,226,213}57 & \cellcolor[RGB]{241,198,173}52 & \cellcolor[RGB]{224,234,246}64 & \cellcolor[RGB]{210,224,242}65 \\
\LogoOpenAI{}GPT-6.1 Sol & \cellcolor{allg}70 & \cellcolor[RGB]{195,213,238}75 & \cellcolor[RGB]{248,227,215}66 & \cellcolor[RGB]{239,189,161}59 & \cellcolor[RGB]{237,178,144}57 & \cellcolor[RGB]{226,235,246}72 & \cellcolor[RGB]{173,198,232}78 & \cellcolor{allg}59 & \cellcolor[RGB]{201,217,239}64 & \cellcolor[RGB]{238,184,153}47 & \cellcolor[RGB]{244,210,190}52 & \cellcolor[RGB]{243,206,184}51 & \cellcolor[RGB]{222,232,245}61 & \cellcolor[RGB]{183,205,234}66 \\
\LogoClaude{}Claude Opus 5.5 & \cellcolor{allg}63 & \cellcolor[RGB]{247,249,252}63 & \cellcolor[RGB]{243,204,182}55 & \cellcolor[RGB]{243,206,185}55 & \cellcolor[RGB]{241,196,170}53 & \cellcolor[RGB]{181,204,234}70 & \cellcolor[RGB]{161,190,228}73 & \cellcolor{allg}59 & \cellcolor[RGB]{252,245,241}58 & \cellcolor[RGB]{238,184,153}48 & \cellcolor[RGB]{247,224,210}55 & \cellcolor[RGB]{249,231,220}56 & \cellcolor[RGB]{189,209,236}66 & \cellcolor[RGB]{166,193,229}69 \\
\LogoOpenAI{}GPT-6.1 Terra & \cellcolor{allg}60 & \cellcolor[RGB]{174,198,232}68 & \cellcolor[RGB]{234,165,125}44 & \cellcolor[RGB]{238,182,150}47 & \cellcolor[RGB]{236,174,138}46 & \cellcolor[RGB]{193,212,237}66 & \cellcolor[RGB]{161,190,228}69 & \cellcolor{allg}60 & \cellcolor[RGB]{210,224,242}64 & \cellcolor[RGB]{241,196,170}50 & \cellcolor[RGB]{250,236,228}57 & \cellcolor[RGB]{245,212,193}53 & \cellcolor[RGB]{245,248,252}61 & \cellcolor[RGB]{201,218,239}65 \\
\LogoGemini{}Gemini 4 Argon & \cellcolor{allg}52 & \cellcolor[RGB]{142,176,222}65 & \cellcolor[RGB]{228,142,92}32 & \cellcolor[RGB]{236,176,141}39 & \cellcolor[RGB]{236,176,141}39 & \cellcolor[RGB]{200,216,239}58 & \cellcolor[RGB]{181,204,234}60 & \cellcolor{allg}54 & \cellcolor[RGB]{178,202,233}62 & \cellcolor[RGB]{239,187,156}43 & \cellcolor[RGB]{239,187,157}43 & \cellcolor[RGB]{250,234,225}51 & \cellcolor[RGB]{220,230,245}57 & \cellcolor[RGB]{224,234,246}57 \\
\LogoOpenAI{}GPT-5.6 Sol & \cellcolor{allg}45 & \cellcolor[RGB]{199,216,239}51 & \cellcolor[RGB]{234,164,124}30 & \cellcolor[RGB]{238,182,149}33 & \cellcolor[RGB]{237,178,145}33 & \cellcolor[RGB]{186,207,235}52 & \cellcolor[RGB]{130,168,219}59 & \cellcolor{allg}49 & \cellcolor[RGB]{206,221,241}53 & \cellcolor[RGB]{237,181,149}36 & \cellcolor[RGB]{247,222,208}44 & \cellcolor[RGB]{236,176,142}35 & \cellcolor[RGB]{200,216,239}54 & \cellcolor[RGB]{178,202,233}56 \\
\LogoClaude{}Claude Haiku 5.5 & \cellcolor{allg}42 & \cellcolor[RGB]{209,223,242}46 & \cellcolor[RGB]{234,165,126}27 & \cellcolor[RGB]{235,170,132}27 & \cellcolor[RGB]{242,202,179}34 & \cellcolor[RGB]{170,196,231}51 & \cellcolor[RGB]{138,174,222}55 & \cellcolor{allg}52 & \cellcolor[RGB]{225,234,246}55 & \cellcolor[RGB]{240,192,165}42 & \cellcolor[RGB]{247,223,210}48 & \cellcolor[RGB]{247,222,207}47 & \cellcolor[RGB]{248,250,253}53 & \cellcolor[RGB]{165,193,229}61 \\
\LogoGemini{}Gemini 3.8 Flash & \cellcolor{allg}40 & \cellcolor[RGB]{192,211,237}46 & \cellcolor[RGB]{237,180,148}28 & \cellcolor[RGB]{235,172,136}26 & \cellcolor[RGB]{250,237,229}38 & \cellcolor[RGB]{216,228,244}44 & \cellcolor[RGB]{175,199,232}48 & \cellcolor{allg}44 & \cellcolor[RGB]{188,208,236}51 & \cellcolor[RGB]{243,206,185}37 & \cellcolor[RGB]{242,199,175}36 & \cellcolor[RGB]{234,240,249}46 & \cellcolor[RGB]{253,249,246}44 & \cellcolor[RGB]{242,246,251}45 \\
\LogoOpenAI{}GPT-6 Luna & \cellcolor{allg}40 & \cellcolor[RGB]{208,222,241}44 & \cellcolor[RGB]{231,152,107}21 & \cellcolor[RGB]{239,186,156}28 & \cellcolor[RGB]{237,179,145}27 & \cellcolor[RGB]{170,196,230}48 & \cellcolor[RGB]{126,165,218}54 & \cellcolor{allg}44 & \cellcolor[RGB]{168,195,230}52 & \cellcolor[RGB]{240,193,166}33 & \cellcolor[RGB]{243,204,182}36 & \cellcolor[RGB]{243,205,183}36 & \cellcolor[RGB]{251,239,233}42 & \cellcolor[RGB]{197,215,238}49 \\
\LogoClaude{}Claude Sonnet 5.5 & \cellcolor{allg}33 & \cellcolor[RGB]{226,235,246}35 & \cellcolor[RGB]{236,174,139}19 & \cellcolor[RGB]{241,194,168}23 & \cellcolor[RGB]{252,246,242}32 & \cellcolor[RGB]{228,236,247}35 & \cellcolor[RGB]{141,176,222}45 & \cellcolor{allg}47 & \cellcolor[RGB]{221,231,245}50 & \cellcolor[RGB]{240,190,162}36 & \cellcolor[RGB]{244,208,188}39 & \cellcolor[RGB]{208,223,241}51 & \cellcolor[RGB]{250,236,228}44 & \cellcolor[RGB]{174,199,232}55 \\
\hline
\LogoKimi{}Kimi-K3.1-Pro & \cellcolor{allg}47 & \cellcolor[RGB]{181,204,234}54 & \cellcolor[RGB]{236,177,142}34 & \cellcolor[RGB]{233,162,120}31 & \cellcolor[RGB]{241,197,171}38 & \cellcolor[RGB]{212,225,243}51 & \cellcolor[RGB]{137,173,221}60 & \cellcolor{allg}45 & \cellcolor[RGB]{215,227,243}48 & \cellcolor[RGB]{246,218,201}39 & \cellcolor[RGB]{243,206,184}37 & \cellcolor[RGB]{250,234,225}42 & \cellcolor[RGB]{253,253,254}45 & \cellcolor[RGB]{193,212,237}51 \\
\LogoDeepSeek{}DeepSeek-V4.1-Flash (1017) & \cellcolor{allg}36 & \cellcolor[RGB]{207,222,241}40 & \cellcolor[RGB]{242,201,177}27 & \cellcolor[RGB]{240,193,166}26 & \cellcolor[RGB]{247,221,205}31 & \cellcolor[RGB]{186,207,235}43 & \cellcolor[RGB]{221,231,245}39 & \cellcolor{allg}45 & \cellcolor[RGB]{180,203,233}53 & \cellcolor[RGB]{242,201,177}37 & \cellcolor[RGB]{240,194,167}35 & \cellcolor[RGB]{248,229,217}42 & \cellcolor[RGB]{224,234,246}48 & \cellcolor[RGB]{241,245,251}46 \\
\LogoMiMo{}MiMo-V2.6-Flash & \cellcolor{allg}35 & \cellcolor[RGB]{195,213,238}41 & \cellcolor[RGB]{238,182,149}23 & \cellcolor[RGB]{238,184,152}23 & \cellcolor[RGB]{248,228,217}32 & \cellcolor[RGB]{222,232,245}38 & \cellcolor[RGB]{173,198,231}44 & \cellcolor{allg}39 & \cellcolor[RGB]{200,217,239}44 & \cellcolor[RGB]{237,178,144}26 & \cellcolor[RGB]{242,200,176}31 & \cellcolor[RGB]{249,251,253}40 & \cellcolor[RGB]{247,249,252}40 & \cellcolor[RGB]{187,208,235}46 \\
\LogoHunyuan{}HY4-preview & \cellcolor{allg}33 & \cellcolor[RGB]{210,224,242}37 & \cellcolor[RGB]{236,176,142}19 & \cellcolor[RGB]{239,187,157}21 & \cellcolor[RGB]{233,164,123}17 & \cellcolor[RGB]{178,202,233}40 & \cellcolor[RGB]{123,163,217}47 & \cellcolor{allg}36 & \cellcolor[RGB]{211,224,242}40 & \cellcolor[RGB]{237,181,148}24 & \cellcolor[RGB]{242,201,178}28 & \cellcolor[RGB]{235,172,135}22 & \cellcolor[RGB]{191,210,236}42 & \cellcolor[RGB]{145,179,224}48 \\
\LogoQwen{}Qwen3.8-27B & \cellcolor{allg}26 & \cellcolor[RGB]{219,230,244}29 & \cellcolor[RGB]{242,202,179}18 & \cellcolor[RGB]{239,187,158}15 & \cellcolor[RGB]{250,237,229}24 & \cellcolor[RGB]{209,223,242}30 & \cellcolor[RGB]{181,204,234}33 & \cellcolor{allg}37 & \cellcolor[RGB]{200,217,239}42 & \cellcolor[RGB]{237,180,147}24 & \cellcolor[RGB]{243,206,185}29 & \cellcolor[RGB]{200,217,239}42 & \cellcolor[RGB]{251,242,236}35 & \cellcolor[RGB]{208,222,241}41 \\
\LogoGLM{}GLM-5.3-Flash & \cellcolor{allg}24 & \cellcolor[RGB]{222,232,245}27 & \cellcolor[RGB]{236,174,139}11 & \cellcolor[RGB]{240,191,163}14 & \cellcolor[RGB]{243,247,251}25 & \cellcolor[RGB]{203,219,240}29 & \cellcolor[RGB]{175,200,232}32 & \cellcolor{allg}34 & \cellcolor[RGB]{193,212,237}40 & \cellcolor[RGB]{231,152,107}15 & \cellcolor[RGB]{241,197,171}24 & \cellcolor[RGB]{155,185,226}44 & \cellcolor[RGB]{251,239,232}32 & \cellcolor[RGB]{200,217,239}39 \\\hline
\end{tabular}}
\end{table*}

\begin{table}[!t]\centering
\caption{\textbf{Run failures.} Share of runs failing in reconstruction and editing, and how each model's failures split by cause.}\label{tab:failures}
\small\setlength{\tabcolsep}{3pt}\renewcommand{\arraystretch}{1.08}
\resizebox{\columnwidth}{!}{%
\begin{tabular}{l|cc|ccc}
 & \multicolumn{2}{c|}{\cellcolor{bandA}Failure rate (\%)} & \multicolumn{3}{c}{\cellcolor{bandB}Cause (\% of failures)} \\
Model & Rec. & Edit & Runtime & Syntax & Timeout \\ \hline
\rowcolor{grp}\multicolumn{6}{l}{\textit{Closed-source Models (API)}} \\
\LogoOpenAI{}GPT-6 Astra & 4.3 & 5.7 & 20 & 30 & 50 \\
\LogoOpenAI{}GPT-6.1 Sol & 5.2 & 6.9 & 35 & 24 & 41 \\
\LogoClaude{}Claude Opus 5.5 & 3.3 & 4.0 & 41 & 0 & 59 \\
\LogoOpenAI{}GPT-6.1 Terra & 4.8 & 5.0 & 43 & 19 & 38 \\
\LogoGemini{}Gemini 4 Argon & 8.3 & 8.6 & 78 & 14 & 8 \\
\LogoOpenAI{}GPT-5.6 Sol & 5.7 & 8.1 & 32 & 3 & 65 \\
\LogoClaude{}Claude Haiku 5.5 & 2.4 & 3.8 & 50 & 0 & 50 \\
\LogoGemini{}Gemini 3.8 Flash & 5.7 & 11.4 & 49 & 6 & 45 \\
\LogoOpenAI{}GPT-6 Luna & 7.1 & 7.6 & 40 & 19 & 41 \\
\LogoClaude{}Claude Sonnet 5.5 & 2.1 & 2.4 & 30 & 40 & 30 \\
\hline
\rowcolor{grp}\multicolumn{6}{l}{\textit{Open-weight Models}} \\
\LogoKimi{}Kimi-K3.1-Pro & 11.0 & 15.0 & 55 & 10 & 35 \\
\LogoDeepSeek{}DeepSeek-V4.1-Flash (1017) & 8.6 & 10.0 & 62 & 5 & 33 \\
\LogoMiMo{}MiMo-V2.6-Flash & 14.5 & 16.4 & 83 & 7 & 10 \\
\LogoHunyuan{}HY4-preview & 36.7 & 37.4 & 40 & 57 & 3 \\
\LogoQwen{}Qwen3.8-27B & 13.1 & 14.5 & 59 & 30 & 11 \\
\LogoGLM{}GLM-5.3-Flash & 11.7 & 13.8 & 79 & 7 & 14 \\
\hline
\end{tabular}}
\end{table}

\begin{figure*}[!t]\centering
\captionsetup{hypcap=false}
\includegraphics[width=\textwidth]{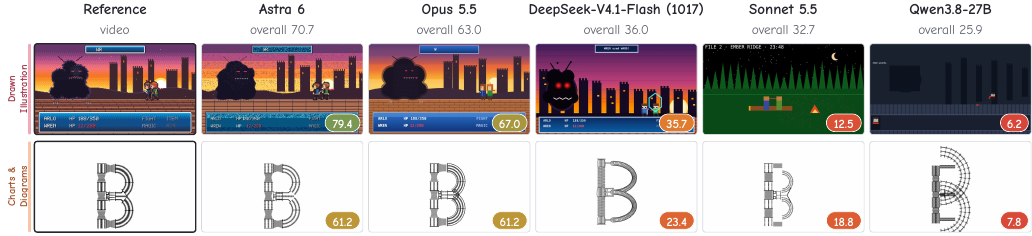}
\caption{\textbf{Reconstruction examples across visual forms.}
Rows show reference and reconstructed frames at matching timestamps. Columns follow overall $\Srecon$; badges show clip scores.}\label{fig:reconcases}

\medskip
\includegraphics[width=\textwidth]{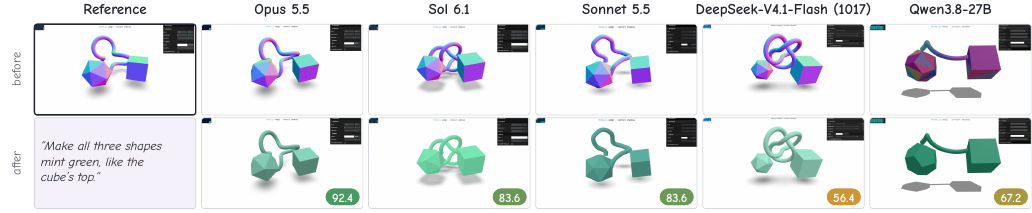}
\caption{\textbf{Editing examples.}
Before and after frames show each model's reconstruction and its edit at the same time. Badges show task scores.
The instruction requests a mint-green recolor.}\label{fig:editcases}
\end{figure*}

\section{Experiments}
\label{sec:exp}

We evaluate 16 code-generating models and 2 diffusion video generation models on \bench; representative prompts are shown in \cref{sec:s_prompts}.

\paragraph{Model choices.}
We select 16 code-generating models spanning multiple model families and access tiers: 10 proprietary API models (across the GPT~\cite{gpt,gpt61}, Claude~\cite{claude}, and Gemini~\cite{gemini} families) and 6 open-weight models (Kimi-K3.1-Pro~\cite{kimi}, DeepSeek-V4.1-Flash (1017)~\cite{deepseek}, MiMo-V2.6-Flash~\cite{mimo}, HY4-preview~\cite{hy4}, Qwen3.8-27B~\cite{qwen}, and GLM-5.3-Flash~\cite{glm}).
As diffusion video generation baselines, MiniMax-H3~\cite{minimax} and Wan2.1-VACE-1.3B~\cite{wan2025,jiang2025vace} edit the reference video directly.
For videos longer than 15\,s, we follow MMLVE-Agent~\cite{wu2026mmlve} and edit them in chunks of at most 15\,s that are then stitched together.

\paragraph{Evaluation metrics.}
Following \cref{sec:eval}, we report $\Srecon$ (combining the binary rubric score $R$ and the holistic rating $H$) for reconstruction, and $\Sedit$ (combining completion $C$, keep rate $K$, and quality $Q$) for editing, averaged across our two VLM judges.
Failed runs score zero.

\paragraph{Inference methods.}
In reconstruction, each model receives only the reference video and returns a single self-contained HTML program.
In editing, the model is given the reference video, its own reconstructed code, and the user instruction, and generates the modified program.
All programs render at the reference's resolution, frame rate, and duration.

\subsection{Quantitative Results}
\label{sec:results}

\paragraph{\bench challenges even frontier models.}
Only 1 out of 16 models (Astra 6) scores above 70 in reconstruction (\cref{tab:recon}) or above 60 in editing (\cref{tab:edit}).
The diffusion-based MiniMax-H3 keeps more of the reference than any code model but completes far fewer requested changes, especially \where and \when ones, so it scores only 39.5, below 12 of the 16 code models; precise local editing still relies on code.
Comparing access tiers also shows a clear gap between closed-source and open-weight models: the top five reconstructors are all proprietary, whereas Kimi-K3.1-Pro leads the open-weight models in 6th place (47.1), outperforming the runner-up open model DeepSeek-V4.1-Flash (1017) by over 11 points, and open models occupy the bottom of the leaderboard.
Examining the rubric dimensions reveals the core bottleneck: across all 16 models, \emph{Motion} is consistently the lowest-scoring dimension, showing that models capture static text, layout, and style far better than motion.

\paragraph{Reconstruction errors carry into editing.}
Comparing \cref{tab:recon,tab:edit}, the two rankings largely agree: Astra 6 leads both stages, and the same four models form the top tier.
Because keep items are checked against the reference video, flaws in a model's own reconstruction persist through its edits, so weak reconstructors also lose keep rate.
The main exceptions are Sonnet 5.5, which climbs from 14th in reconstruction to 8th in editing, and Kimi-K3.1-Pro, which falls from 6th to 10th.
Across edit categories, spatial (\where) and appearance (\how) changes are completed reasonably well, whereas temporal (\when) changes are hardest for all.

\begin{table}[!t]\centering
\caption{\textbf{Rendering techniques.} Share (\%) of each model's reconstructions by primary technique.}\label{tab:techstack}
\small\setlength{\tabcolsep}{3.2pt}\renewcommand{\arraystretch}{1.08}
\resizebox{\columnwidth}{!}{%
\begin{tabular}{l|ccccc}
Model & \makecell{Canvas\\2D} & SVG & \makecell{Three\\.js} & WebGL & \makecell{HTML\\/CSS} \\ \hline
\LogoOpenAI{}GPT-6 Astra & 83 & 6 & 8 & 2 & 0 \\
\LogoOpenAI{}GPT-6.1 Sol & 79 & 9 & 9 & 2 & 0 \\
\LogoClaude{}Claude Opus 5.5 & 82 & 7 & 9 & 2 & 0 \\
\LogoOpenAI{}GPT-6.1 Terra & 82 & 9 & 7 & 2 & 0 \\
\LogoGemini{}Gemini 4 Argon & 79 & 4 & 11 & 4 & 2 \\
\LogoOpenAI{}GPT-5.6 Sol & 70 & 20 & 8 & 1 & 2 \\
\LogoClaude{}Claude Haiku 5.5 & 78 & 14 & 5 & 1 & 2 \\
\LogoGemini{}Gemini 3.8 Flash & 60 & 22 & 13 & 5 & 1 \\
\LogoOpenAI{}GPT-6 Luna & 56 & 35 & 5 & 1 & 1 \\
\LogoClaude{}Claude Sonnet 5.5 & 86 & 6 & 3 & 1 & 4 \\
\hline
\LogoKimi{}Kimi-K3.1-Pro & 79 & 9 & 10 & 2 & 0 \\
\LogoDeepSeek{}DeepSeek-V4.1-Flash (1017) & 81 & 8 & 9 & 2 & 1 \\
\LogoMiMo{}MiMo-V2.6-Flash & 72 & 14 & 11 & 2 & 1 \\
\LogoHunyuan{}HY4-preview & 87 & 3 & 3 & 7 & 0 \\
\LogoQwen{}Qwen3.8-27B & 82 & 5 & 10 & 1 & 2 \\
\LogoGLM{}GLM-5.3-Flash & 79 & 6 & 12 & 2 & 2 \\
\hline
\end{tabular}}
\end{table}

\begin{table}[t]\centering
\caption{\textbf{Correlation between judge and human scores}, computed per annotator and averaged over three annotators; judge scores average both judges.}\label{tab:human}
\small
\begin{tabular*}{\columnwidth}{@{\extracolsep{\fill}}llcc}
\toprule
Task & Dimension & Pearson & Spearman \\
\midrule
\multirow{2}{*}{Reconstruction} & Rubric ($R$) & 0.86 & 0.85 \\
 & Holistic ($H$) & 0.84 & 0.86 \\
\midrule
\multirow{3}{*}{Editing} & Completion ($C$) & 0.87 & 0.88 \\
 & Keep rate ($K$) & 0.81 & 0.80 \\
 & Quality ($Q$) & 0.84 & 0.84 \\
\bottomrule
\end{tabular*}
\end{table}

\paragraph{Performance varies across visual forms.}
Dividing results across visual forms in \cref{tab:bycat} shows a consistent gap between structured and freeform visuals.
Highly structured formats such as charts and UI animations are the easiest to reconstruct, with charts ranking as the easiest form for 14 of the 16 models, as their regular layouts map naturally to standard web elements.
In contrast, scores fall by up to about 20 points below each model's average on hand-drawn illustrations, 3D scenes, and particle effects.
These forms require complex geometry or organic motion, which is hard to express with simple layout rules.
Editing shows the same pattern with smaller gaps, likely because an edit changes only a few elements of an existing program rather than rebuilding the whole scene.
Hand-drawn illustrations remain the hardest form to edit for 11 of the 16 models, whereas particle effects come much closer to the average.

\paragraph{Generating executable animation code remains fragile, and models default to low-level 2D drawing over structured graphics.}
Beyond visual fidelity, models differ sharply in whether their generated programs can even execute and render properly (\cref{tab:failures}).
While the Claude family fails on at most 4\% of runs in either stage, open-weight models fail on 9--37\% of runs.
Rather than simple syntax mistakes, most failures stem from rendering timeouts and silent runtime errors that produce blank frames, showing that models struggle to manage complex animation loops and execution state.
When programs do run, models exhibit a strong paradigm bias toward imperative Canvas 2D scripts, using them in 56--87\% of all reconstructions (\cref{tab:techstack}).
Instead of selecting declarative SVG for vector graphics or Three.js (at most 13\%) for 3D scenes, models consistently fall back on procedural 2D pixel drawing regardless of the input.

\subsection{Qualitative Analysis}
\label{sec:qual}

\paragraph{Reconstruction.}
Visual examples in \cref{fig:reconcases} show that weaker models often recognize coarse semantics but lose fine structure.
On the \emph{Charts \& Diagrams} clip, Astra 6 and Opus 5.5 closely reproduce the letter B with its multi-layered strokes, whereas Qwen3.8-27B breaks the overall structure so badly that the letter is barely recognizable.

\paragraph{Editing.}
Asked to recolor all three 3D shapes mint green (\cref{fig:editcases}), Opus 5.5 and Sol 6.1 apply the new color while keeping the original shading and geometry intact.
DeepSeek-V4.1-Flash (1017) and Qwen3.8-27B both edit reconstructions whose structure is already wrong, and these errors carry into their results.
Qwen3.8-27B also renders a dark green far from the requested mint.
Keep rate and quality capture these failures, which a color check alone would miss.

\subsection{Reliability of the Evaluation}
\label{sec:reliability}

Our two VLM judges show strong consistency, ordering 97\% of model pairs identically in reconstruction and 92\% in editing.
Frame-similarity metrics, by contrast, compress all models into a narrow band and even misorder some of them (\cref{tab:recon}), as they mainly reflect palette and layout rather than motion and timing.
To validate our automatic protocol against human perception, three annotators independently rated 500 random items, each a model's reconstruction and its edit of one clip, on the same five scores ($R$, $H$, $C$, $K$, $Q$).
As shown in \cref{tab:human}, our rubric-based scores align closely with human judgments, achieving Pearson correlations of 0.81--0.87.

\section{Conclusion}
\label{sec:conclusion}
We presented \bench, a benchmark that evaluates whether frontier LLMs can reconstruct a reference video as an executable program from pixels alone and then edit their own code following natural-language instructions.
Organized under a unified \where/\when/\how taxonomy, \bench pairs realistic recreate-then-edit workflows with a decomposed evaluation protocol that separates missed edits, wrong edits, and collateral damage without requiring ground-truth edited videos.
Benchmarking 16 code-generating models and 2 diffusion video generation models shows that while code enables more precise local edits than direct video generation, reconstruction errors carry into editing, motion and timing remain the main bottleneck, and even the best editors damage content they should preserve.
We hope \bench drives progress in temporally grounded visual programming and self-editing agents.

\section*{LLM Usage Statement}
LLMs are part of this research: the evaluated models are LLMs and VLMs, and Sol 6.1 and Sonnet 5.5 serve as judges (\cref{sec:eval}).
We also used LLM assistants for code assistance, grammar checking, and manuscript polishing.
The authors take full responsibility for the code, manuscript, and all scientific claims.

{
    \small
    \bibliographystyle{ieeenat_fullname}
    \bibliography{main}
}

\setcounter{figure}{0}
\setcounter{table}{0}
\renewcommand{\thetable}{A\arabic{table}}
\renewcommand{\thefigure}{A\arabic{figure}}
\twocolumn[{\centering\Large\textbf{Appendix}\par\vspace{1em}}]
\appendix
\noindent This appendix gives related work (\cref{sec:s_related}), benchmark construction (\cref{sec:s_data}), evaluation details (\cref{sec:s_eval}), detailed results (\cref{sec:s_results}), judge agreement (\cref{sec:s_rel}), the prompts (\cref{sec:s_prompts}), and qualitative comparisons of the models (\cref{sec:s_cases}).

\section{Related Work}
\label{sec:s_related}

\paragraph{Code-based visual generation.}
Proprietary models such as GPT~\cite{gpt}, Claude~\cite{claude}, and Gemini~\cite{gemini}, and open-weight models such as Qwen~\cite{qwen} and Kimi~\cite{kimi}, now both understand images and write code, which has made code-based generation of visual content an active direction of multimodal research.
StarVector~\cite{rodriguez2025starvector} generates SVG code directly from images.
For dynamic content, Code2Video~\cite{chen2025code2video} and TheoremExplainAgent~\cite{ku2025theoremexplainagent} produce educational videos by writing Manim code, while LogoMotion~\cite{liu2024logomotion} and LottieGPT~\cite{lottiegpt} generate logo animations and vector animations.
Videos produced this way are rendered by software rather than captured by a camera or sampled from a diffusion model, so every element and its motion are specified exactly by the program and remain editable.
These methods start from text or static images; \bench instead starts from a reference video, which the model must reproduce as a program and then edit.

\paragraph{Benchmarks for code-based visual generation and editing.}
\looseness=-1 ChartMimic~\cite{yang2025chartmimic} evaluates generating plotting code from chart images, and Design2Code~\cite{si2024design2code} and DesignBench~\cite{designbench} turn webpage screenshots into front-end code.
Animation2Code~\cite{ji2026animation2code}, WebVR~\cite{dai2026webvr}, and IWR-Bench~\cite{iwrbench} extend the input to videos of web animations and interactions; Animation2Code finds that temporal fidelity lags far behind appearance fidelity, and IWR-Bench that functional correctness lags far behind visual fidelity.
Code2Video-Compass~\cite{chen2026compass} builds a benchmark for code-based video generation from text, image, and video conditions, and BVB~\cite{tang2026bvb}, 4DCodeBench~\cite{shen2026fourdcodebench}, and CineRecon~\cite{cinerecon} reconstruct dynamic 3D scenes as Blender programs.
A separate line of work evaluates editing existing programs: Chart2Code~\cite{tang2025chart2code}, ChartEdit~\cite{zhao2025chartedit}, and RealChart2Code~\cite{zhang2026realchart2code} add chart editing on top of reproduction, BlenderGym~\cite{gu2025blendergym} studies 3D scene editing, and EDIT-Bench~\cite{editbench2025} targets code edits in real development.
PPTArena~\cite{ofengenden2025pptarena} edits real slide decks, including their animations, but operates on slide decks rather than videos.
These editing benchmarks start from given source code and mostly edit static content, while video-to-code benchmarks stop at reconstruction.
\bench connects the two lines: the model writes a program from the reference video alone and then edits it, including temporal edits.

\section{Benchmark Construction}
\label{sec:s_data}

\paragraph{Sources and filtering.}
The 420 clips come from Lottie (162 clips), LemoLab (74), web repositories (60), YouTube and other video sites (87), and X (37).
Lottie (After Effects keyframe JSON played by lottie-web), LemoLab (React/Remotion), and the web repositories (HTML, Canvas, and WebGL) ship with code; YouTube and X clips are video only.
The code is used only to write and check editing tasks; no model under test sees it.
From about 5.8K candidates, rule filters keep clips of 3\,s to 3\,min with at least three moving regions and, for sources with code, at least 30 lines of code; they also remove near duplicates and keep at most five clips per author.
Sol 6.1 then drops static, live-footage, or too simple clips.

\begin{figure*}[t]
  \centering
  \includegraphics[width=0.82\textwidth]{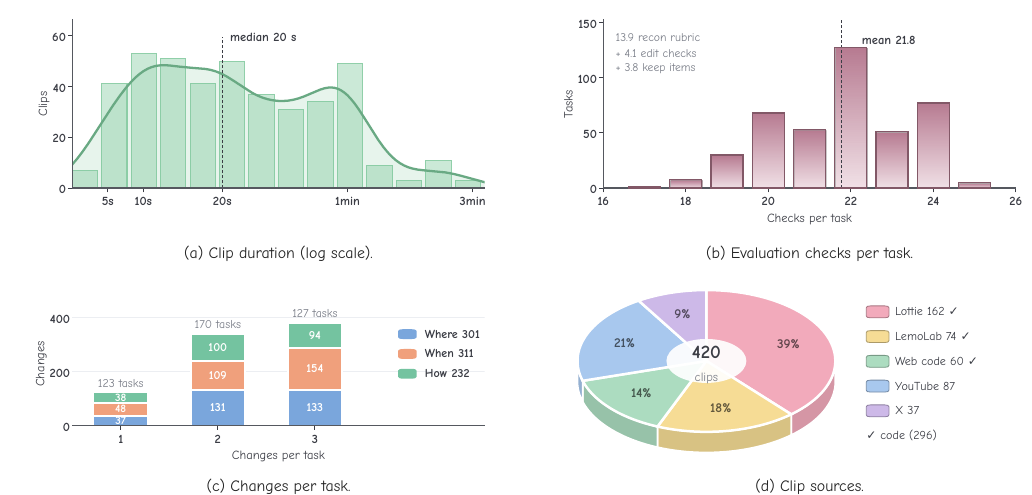}
  \caption{\textbf{Further dataset statistics.}
  (a) Clip duration (log scale); (b) evaluation checks per task; (c) changes per task, split into \where, \when, and \how; (d) clip sources, where \ding{51} marks sources that come with code.}
  \label{fig:supp_stats}
\end{figure*}

\paragraph{Statistics.}
\looseness=-1 Each clip carries one content type: Explainer (112 clips), Narrative (83), Brand Promo (75), Abstract (64), Product Demo (57), and Charts \& Data (29).
Durations range from 2.9\,s to 179.3\,s (median 20\,s), and each clip has 21.8 evaluation criteria on average (13.9 rubric items, 4.1 edit checks, and 3.8 keep items), with further distributions in \cref{fig:supp_stats}.

\paragraph{Editing tasks.}
\label{sec:s_tasks}
Each requirement belongs to one domain: \where (position, size, anchor, crop, padding, clipping, stacking, and layout), \when (time placement, cycling, trimming, entrance, sustained motion, exit, transitions, and fades), or \how (text, typography, fill and stroke, image adjustment, shadow, and effects).
The 420 instructions hold 844 requirements (301 \where, 311 \when, 232 \how); 123, 170, and 127 instructions contain one, two, and three requirements.
Sol 6.1 writes each instruction from a sampled slot (number and type of changes, phrasing, and time region), seeing the author code when there is one so that requests are feasible, and uses only visible terms, never variable or layer names.
Each requirement gets a time window and one to three yes/no checks, and each task gets two to four keep items.
An annotator checked every task: each requirement is visible and feasible, checks and keep items match the instruction, and no code identifiers appear.

\section{Evaluation Details}
\label{sec:s_eval}

\paragraph{Rendering and frames.}
\looseness=-1 Each output is one HTML page, rendered in headless Chromium under a virtual clock from $t{=}0$.
Code models receive frames of the reference video sampled at 1\,fps (long side 768\,px).
The reconstruction judge compares reference and candidate frames at the same timestamps, sampled at 1\,fps.
The edit judge compares reference and edited frames at the same moments, taken inside each requested change's time window ($\pm$0.5\,s) and evenly over the whole clip.

\paragraph{Scoring.}
\looseness=-1 A rubric dimension passes only if all of its items pass, and $R$ is the weighted pass rate over dimensions; the four holistic ratings are mapped to $[0,1]$ and averaged into $H$.
Each edit check is judged done (1), partial (0.5), or missed (0), and $C$ averages checks within each requirement and then over requirements; if nothing visibly changes in a requirement's window, all its checks count as missed.
Failed runs (no page, a render error, or a timeout) score zero.

\FloatBarrier
\section{Additional Results}
\label{sec:s_results}

\begin{figure}[t]
  \centering
  \includegraphics[width=\linewidth]{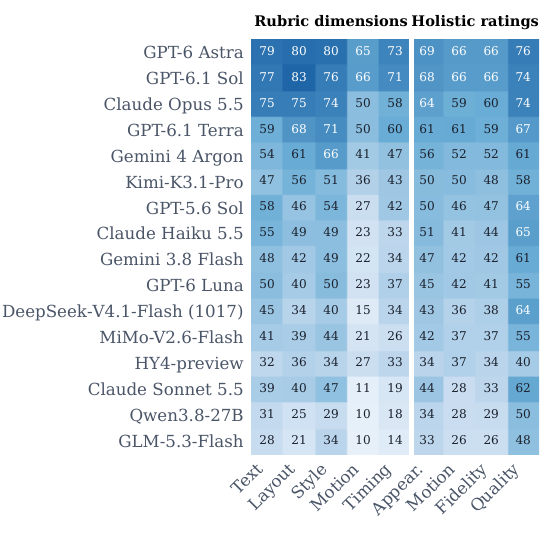}
  \caption{\textbf{Reconstruction scores per dimension.} Motion and timing are the weakest columns for nearly every model.}
  \label{fig:s_recondim}
\end{figure}

\begin{figure}[t]
  \centering
  \includegraphics[width=\linewidth]{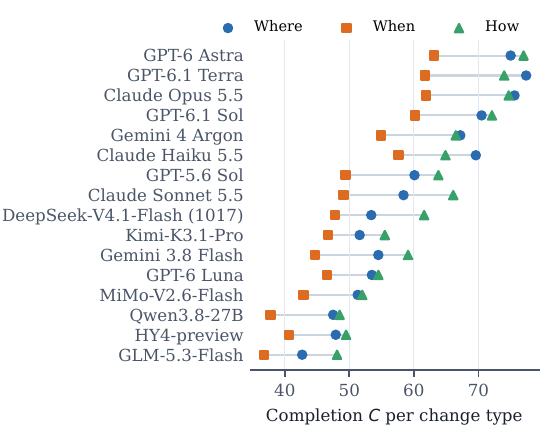}
  \caption{\textbf{Completion per change type}, with models sorted by $\Sedit$. \when (orange squares) trails for every model.}
  \label{fig:s_wwh}
\end{figure}

\begin{figure*}[tp]
  \centering
  \includegraphics[width=0.7\textwidth]{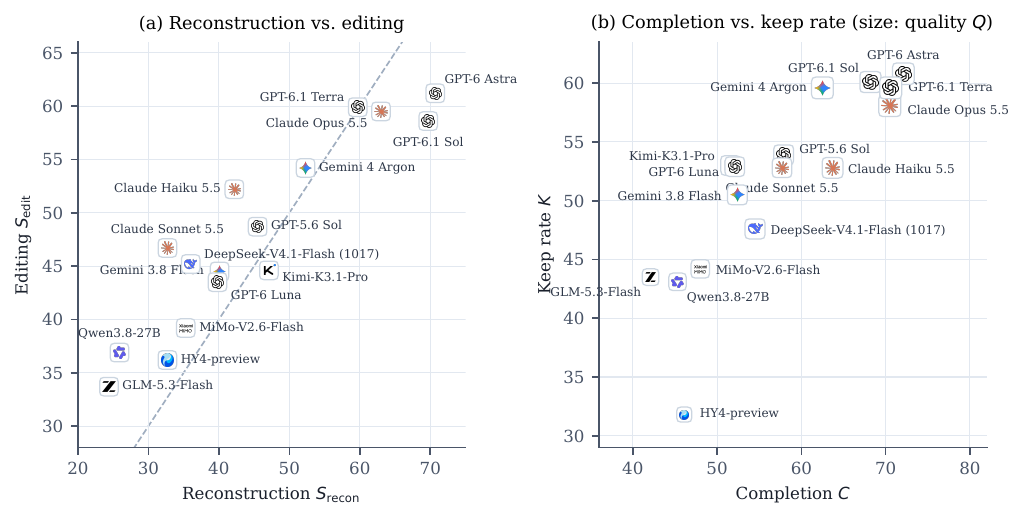}
  \caption{\textbf{Model-level comparisons.} (a) Reconstruction against editing score; the dashed line marks equal scores. (b) Completion against keep rate, with marker size proportional to quality.}
  \label{fig:s_scatter}
\end{figure*}

\begin{figure*}[tp]
  \centering
  \includegraphics[width=0.7\textwidth]{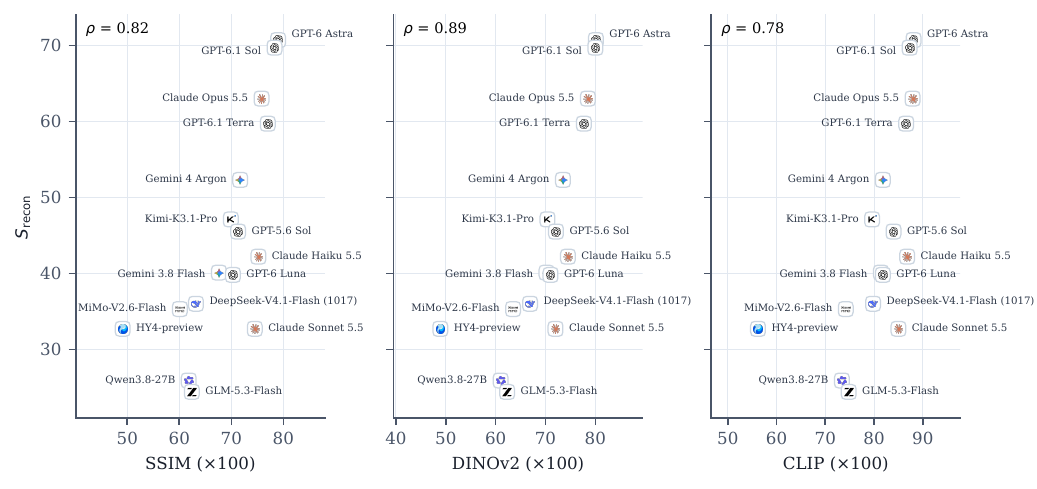}
  \caption{\textbf{Objective metrics against $\Srecon$.} $\rho$: Spearman correlation over the 16 models.}
  \label{fig:s_objective}
\end{figure*}

\noindent Reconstruction by rubric dimension (\cref{tab:s_recondim,fig:s_recondim}), editing by change type (\cref{tab:s_editwwh,fig:s_wwh}), scores by content type and visual form (\cref{tab:s_content,tab:s_form}), reconstruction against editing (\cref{fig:s_scatter}), video generation baselines (\cref{tab:s_video}), and objective metrics (\cref{fig:s_objective,tab:s_obj}).

\looseness=-1 Motion is the weakest dimension for all 16 models and timing the second weakest for 14, while text, layout, and style are much closer to solved for the strongest models; holistic quality stays high even for weak models, as it rates rendering artifacts, not fidelity to the reference.
\when changes have the lowest completion for every code model.
Narrative clips are the hardest content type to rebuild and edit for most models, and charts are among the easiest.
Models that rebuild well also tend to edit well (\cref{fig:s_scatter}a), but some sit well above the trend: Claude Sonnet 5.5 ranks 14th in reconstruction but 8th in editing.
Frame-similarity metrics rank models broadly as the judges do (\cref{fig:s_objective}), but Claude Sonnet 5.5 and Claude Haiku 5.5 show their limits: matching palettes and layouts put them in the top six on SSIM and CLIP, while the judges penalize their motion and timing.

\begin{table*}[t]
  \centering
  \small
  \setlength{\tabcolsep}{4pt}
  \caption{\textbf{Reconstruction by dimension.}}
  \label{tab:s_recondim}
  \begin{tabular}{@{}lc@{\hspace{10pt}}cccccc@{\hspace{10pt}}ccccc@{}}
    \toprule
    & & \multicolumn{6}{c}{Rubric} & \multicolumn{5}{c}{Holistic} \\
    \cmidrule(lr){3-8}\cmidrule(l){9-13}
    Model & $\Srecon$ & $R$ & Text & Layout & Style & Motion & Timing & $H$ & Appear. & Motion & Fidelity & Quality \\
    \midrule
GPT-6 Astra & \textbf{70.7} & \textbf{72.2} & \textbf{79.0} & \underline{80.5} & \textbf{79.5} & \underline{65.2} & \textbf{73.4} & \textbf{69.2} & \textbf{68.8} & \textbf{66.1} & \textbf{66.3} & \textbf{75.6} \\
GPT-6.1 Sol & \underline{69.7} & \underline{71.2} & \underline{77.0} & \textbf{82.7} & \underline{75.6} & \textbf{65.5} & \underline{71.3} & \underline{68.2} & \underline{67.5} & \underline{65.5} & \underline{65.5} & \underline{74.5} \\
Claude Opus 5.5 & 63.0 & 61.5 & 75.4 & 75.4 & 74.5 & 49.5 & 58.0 & 64.5 & 64.2 & 58.9 & 60.4 & \underline{74.5} \\
GPT-6.1 Terra & 59.7 & 57.4 & 59.1 & 67.9 & 70.7 & 49.6 & 60.2 & 61.9 & 60.6 & 61.1 & 59.3 & 66.6 \\
Gemini 4 Argon & 52.3 & 49.2 & 53.7 & 60.8 & 65.9 & 41.3 & 47.0 & 55.5 & 56.2 & 52.2 & 52.3 & 61.2 \\
Kimi-K3.1-Pro & 47.1 & 42.4 & 47.3 & 56.0 & 51.2 & 36.3 & 42.6 & 51.7 & 50.4 & 50.1 & 48.5 & 57.7 \\
GPT-5.6 Sol & 45.5 & 39.3 & 57.8 & 45.7 & 54.3 & 27.3 & 42.3 & 51.7 & 50.1 & 46.2 & 46.6 & 63.7 \\
Claude Haiku 5.5 & 42.2 & 34.2 & 54.9 & 48.6 & 48.9 & 22.6 & 32.7 & 50.2 & 50.8 & 41.2 & 43.8 & 64.9 \\
Gemini 3.8 Flash & 40.1 & 32.3 & 47.9 & 42.0 & 49.1 & 21.5 & 33.7 & 47.9 & 46.8 & 41.5 & 42.5 & 60.8 \\
GPT-6 Luna & 39.8 & 33.8 & 49.7 & 40.3 & 50.5 & 22.7 & 37.0 & 45.8 & 44.6 & 42.0 & 41.3 & 55.3 \\
DeepSeek-V4.1-Flash (1017) & 36.0 & 26.8 & 44.7 & 34.5 & 40.1 & 15.4 & 33.8 & 45.2 & 42.8 & 35.8 & 37.7 & 64.4 \\
MiMo-V2.6-Flash & 35.3 & 28.2 & 40.6 & 39.2 & 44.5 & 21.0 & 26.3 & 42.5 & 41.5 & 36.9 & 36.7 & 54.9 \\
HY4-preview & 32.7 & 29.1 & 32.5 & 36.1 & 34.5 & 26.6 & 32.9 & 36.3 & 34.4 & 36.6 & 34.0 & 40.3 \\
Claude Sonnet 5.5 & 32.7 & 23.9 & 39.3 & 39.9 & 47.2 & 10.6 & 19.0 & 41.5 & 44.0 & 27.5 & 32.8 & 61.6 \\
Qwen3.8-27B & 25.9 & 16.9 & 31.2 & 24.6 & 29.1 & 10.5 & 18.5 & 35.0 & 33.7 & 27.7 & 28.8 & 49.6 \\
GLM-5.3-Flash & 24.4 & 15.5 & 28.4 & 20.6 & 34.0 & 9.8 & 14.2 & 33.2 & 32.9 & 26.0 & 26.0 & 48.1 \\
    \bottomrule
  \end{tabular}
\end{table*}

\begin{table*}[t]
  \centering
  \small
  \setlength{\tabcolsep}{4.2pt}
  \caption{\textbf{Editing by change type.}}
  \label{tab:s_editwwh}
  \begin{tabular}{@{}lc@{\hspace{10pt}}ccc@{\hspace{8pt}}ccc@{\hspace{8pt}}ccc@{}}
    \toprule
    & & \multicolumn{3}{c}{\where} & \multicolumn{3}{c}{\when} & \multicolumn{3}{c}{\how} \\
    \cmidrule(lr){3-5}\cmidrule(lr){6-8}\cmidrule(l){9-11}
    Model & $\Sedit$ & $C$ & $K$ & $Q$ & $C$ & $K$ & $Q$ & $C$ & $K$ & $Q$ \\
    \midrule
GPT-6 Astra & \textbf{61.2} & 75.0 & 60.7 & 50.5 & \textbf{63.1} & \textbf{64.0} & 52.5 & \textbf{77.0} & \textbf{59.2} & 51.6 \\
GPT-6.1 Sol & 58.6 & 70.5 & 60.4 & 48.7 & 60.2 & \underline{62.8} & 51.2 & 72.1 & 56.8 & 49.4 \\
Claude Opus 5.5 & 59.5 & \underline{75.6} & 60.1 & \textbf{53.6} & \underline{61.9} & 60.5 & \textbf{53.6} & \underline{74.7} & 56.4 & \underline{53.1} \\
GPT-6.1 Terra & \underline{59.9} & \textbf{77.4} & \textbf{61.8} & \underline{52.8} & 61.7 & 62.1 & \underline{53.0} & 74.0 & \textbf{59.2} & \textbf{54.1} \\
Gemini 4 Argon & 54.2 & 67.2 & \underline{60.8} & 51.5 & 54.9 & 61.4 & 52.6 & 66.5 & \underline{58.4} & 52.1 \\
Kimi-K3.1-Pro & 44.6 & 51.6 & 51.5 & 43.5 & 46.7 & 56.2 & 46.8 & 55.5 & 51.0 & 45.6 \\
GPT-5.6 Sol & 48.7 & 60.1 & 55.9 & 46.4 & 49.4 & 55.5 & 45.5 & 63.8 & 53.4 & 48.1 \\
Claude Haiku 5.5 & 52.2 & 69.6 & 54.2 & 48.9 & 57.6 & 55.9 & 48.6 & 64.9 & 51.5 & 49.9 \\
Gemini 3.8 Flash & 44.5 & 54.5 & 50.7 & 46.4 & 44.7 & 52.5 & 45.7 & 59.1 & 51.2 & 47.2 \\
GPT-6 Luna & 43.5 & 53.5 & 53.9 & 44.5 & 46.5 & 55.3 & 44.3 & 54.5 & 52.7 & 46.3 \\
DeepSeek-V4.1-Flash (1017) & 45.2 & 53.4 & 49.6 & 45.8 & 47.8 & 50.1 & 47.2 & 61.6 & 46.3 & 46.5 \\
MiMo-V2.6-Flash & 39.2 & 51.3 & 46.9 & 42.0 & 42.9 & 45.1 & 41.3 & 52.0 & 42.3 & 41.4 \\
HY4-preview & 36.2 & 47.9 & 33.9 & 31.7 & 40.7 & 34.4 & 32.1 & 49.5 & 30.9 & 32.2 \\
Claude Sonnet 5.5 & 46.7 & 58.4 & 56.3 & 45.5 & 49.1 & 54.6 & 45.8 & 66.1 & 50.2 & 46.3 \\
Qwen3.8-27B & 36.9 & 47.5 & 43.8 & 38.1 & 37.8 & 43.6 & 37.0 & 48.5 & 42.3 & 39.8 \\
GLM-5.3-Flash & 33.7 & 42.7 & 45.2 & 34.7 & 36.8 & 43.7 & 35.2 & 48.1 & 44.1 & 38.3 \\
    \bottomrule
  \end{tabular}
\end{table*}

\begin{table*}[t]
  \centering
  \small
  \setlength{\tabcolsep}{4pt}
  \caption{\textbf{Scores by content type.}}
  \label{tab:s_content}
  \begin{tabular}{@{}lcccccc@{\hspace{12pt}}cccccc@{}}
    \toprule
    & \multicolumn{6}{c}{Reconstruction $\Srecon$} & \multicolumn{6}{c}{Editing $\Sedit$} \\
    \cmidrule(lr){2-7}\cmidrule(l){8-13}
    Model & Expl. & Charts & Narr. & Brand & Product & Abstr. & Expl. & Charts & Narr. & Brand & Product & Abstr. \\
    & (112) & (29) & (83) & (75) & (57) & (64) & (112) & (29) & (83) & (75) & (57) & (64) \\
    \midrule
GPT-6 Astra & \underline{74.0} & \textbf{74.2} & \underline{65.6} & \textbf{71.6} & \textbf{78.2} & \textbf{62.2} & \underline{60.8} & \textbf{66.8} & \underline{52.3} & \textbf{64.1} & \underline{66.3} & \textbf{62.9} \\
GPT-6.1 Sol & \textbf{74.6} & \underline{71.0} & \textbf{67.4} & \underline{69.0} & \underline{76.1} & \underline{58.7} & \textbf{60.9} & 64.8 & 47.6 & 59.7 & \textbf{67.6} & 56.9 \\
Claude Opus 5.5 & 67.0 & 69.8 & 54.7 & 62.0 & 72.5 & 56.3 & 60.4 & 65.0 & 50.5 & 59.8 & 64.3 & \underline{62.5} \\
GPT-6.1 Terra & 62.2 & 63.8 & 51.4 & 61.2 & 72.7 & 50.6 & 60.4 & 64.5 & \textbf{53.1} & \underline{61.9} & 64.6 & 59.0 \\
Gemini 4 Argon & 55.2 & 57.7 & 41.8 & 55.7 & 63.6 & 44.5 & 53.2 & \underline{65.4} & 47.3 & 57.5 & 57.3 & 53.5 \\
Kimi-K3.1-Pro & 47.7 & 57.1 & 36.0 & 52.2 & 56.3 & 41.5 & 39.2 & 49.6 & 40.6 & 52.1 & 45.5 & 47.1 \\
GPT-5.6 Sol & 49.2 & 54.5 & 34.9 & 46.5 & 56.0 & 37.9 & 46.2 & 51.2 & 39.8 & 55.8 & 55.7 & 49.0 \\
Claude Haiku 5.5 & 44.9 & 50.1 & 28.5 & 46.2 & 56.5 & 34.0 & 51.7 & 56.9 & 46.0 & 56.0 & 56.1 & 51.1 \\
Gemini 3.8 Flash & 42.4 & 40.2 & 31.1 & 42.3 & 49.6 & 36.5 & 44.5 & 42.1 & 40.0 & 46.6 & 49.1 & 44.9 \\
GPT-6 Luna & 43.5 & 50.7 & 28.3 & 40.7 & 51.1 & 32.1 & 41.6 & 49.1 & 42.2 & 46.2 & 46.9 & 40.1 \\
DeepSeek-V4.1-Flash (1017) & 36.0 & 42.6 & 28.0 & 42.9 & 42.2 & 29.9 & 40.0 & 51.0 & 39.0 & 52.0 & 49.7 & 47.9 \\
MiMo-V2.6-Flash & 35.6 & 44.1 & 23.1 & 42.9 & 41.8 & 32.2 & 35.6 & 43.9 & 29.3 & 48.4 & 40.7 & 44.1 \\
HY4-preview & 39.4 & 29.8 & 23.2 & 33.6 & 38.4 & 28.8 & 39.1 & 34.2 & 26.1 & 41.6 & 42.1 & 33.6 \\
Claude Sonnet 5.5 & 34.4 & 43.9 & 20.9 & 36.3 & 38.1 & 31.0 & 45.3 & 53.8 & 39.2 & 49.3 & 50.0 & 49.4 \\
Qwen3.8-27B & 26.2 & 37.6 & 17.8 & 29.6 & 32.2 & 21.0 & 31.5 & 46.3 & 29.5 & 41.6 & 43.3 & 40.4 \\
GLM-5.3-Flash & 25.0 & 31.3 & 15.3 & 27.2 & 29.8 & 23.8 & 29.7 & 43.6 & 23.6 & 39.6 & 35.5 & 41.1 \\
    \bottomrule
  \end{tabular}
\end{table*}

\begin{table*}[t]
  \centering
  \small
  \setlength{\tabcolsep}{3.6pt}
  \caption{\textbf{Scores by visual form.}}
  \label{tab:s_form}
  \begin{tabular}{@{}lcccccc@{\hspace{10pt}}cccccc@{}}
    \toprule
    & \multicolumn{6}{c}{Reconstruction $\Srecon$} & \multicolumn{6}{c}{Editing $\Sedit$} \\
    \cmidrule(lr){2-7}\cmidrule(l){8-13}
    Model & Flat & Hand & 3D & Particle & UI & Chart & Flat & Hand & 3D & Particle & UI & Chart \\
    & (123) & (46) & (62) & (46) & (75) & (68) & (123) & (46) & (62) & (46) & (75) & (68) \\
    \midrule
GPT-6 Astra & \textbf{76.2} & \underline{64.4} & \textbf{61.5} & \textbf{58.0} & \textbf{74.4} & \underline{77.9} & \textbf{68.8} & 45.2 & \underline{57.0} & 52.1 & \underline{63.8} & 65.2 \\
GPT-6.1 Sol & \underline{75.4} & \textbf{65.8} & \underline{58.9} & \underline{56.6} & \underline{72.2} & \textbf{78.0} & 63.7 & 46.8 & 51.7 & 50.9 & 61.5 & \underline{65.7} \\
Claude Opus 5.5 & 63.5 & 55.0 & 55.4 & 53.4 & 70.3 & 72.8 & 58.4 & \underline{47.6} & 55.0 & \textbf{56.1} & \textbf{65.9} & \textbf{68.6} \\
GPT-6.1 Terra & 67.9 & 44.0 & 47.5 & 45.8 & 65.6 & 69.4 & \underline{63.9} & \textbf{50.3} & \textbf{57.5} & \underline{53.3} & 60.6 & 64.9 \\
Gemini 4 Argon & 64.5 & 31.8 & 38.9 & 38.9 & 57.5 & 59.7 & 61.9 & 42.9 & 42.9 & 51.5 & 57.3 & 56.8 \\
Kimi-K3.1-Pro & 54.4 & 33.7 & 30.7 & 37.7 & 50.9 & 59.9 & 48.1 & 39.1 & 36.9 & 41.8 & 44.6 & 50.5 \\
GPT-5.6 Sol & 50.8 & 29.7 & 33.1 & 32.5 & 52.2 & 59.2 & 53.3 & 36.4 & 44.0 & 35.4 & 54.0 & 56.4 \\
Claude Haiku 5.5 & 46.3 & 26.6 & 27.5 & 33.9 & 50.9 & 54.8 & 54.7 & 42.0 & 47.7 & 47.4 & 52.7 & 61.5 \\
Gemini 3.8 Flash & 46.2 & 27.5 & 25.9 & 37.7 & 43.5 & 48.2 & 51.1 & 36.9 & 35.6 & 46.2 & 43.9 & 45.4 \\
GPT-6 Luna & 44.1 & 21.4 & 28.3 & 26.9 & 48.5 & 54.0 & 52.4 & 33.5 & 35.6 & 35.7 & 41.6 & 49.0 \\
DeepSeek-V4.1-Flash (1017) & 40.4 & 27.5 & 26.0 & 31.0 & 42.8 & 38.9 & 52.7 & 36.6 & 35.3 & 41.6 & 47.8 & 46.2 \\
MiMo-V2.6-Flash & 41.1 & 23.0 & 23.4 & 31.6 & 38.1 & 43.6 & 44.4 & 26.2 & 30.5 & 39.5 & 39.7 & 45.8 \\
HY4-preview & 36.8 & 19.3 & 21.4 & 16.7 & 40.4 & 47.3 & 40.2 & 23.7 & 27.7 & 22.0 & 42.4 & 47.9 \\
Claude Sonnet 5.5 & 35.1 & 18.9 & 22.9 & 31.7 & 34.9 & 45.0 & 49.6 & 36.1 & 39.5 & 50.9 & 44.3 & 54.8 \\
Qwen3.8-27B & 29.0 & 17.6 & 14.8 & 23.6 & 30.1 & 33.2 & 42.1 & 24.3 & 29.3 & 42.1 & 35.3 & 41.2 \\
GLM-5.3-Flash & 27.1 & 10.6 & 13.9 & 25.2 & 29.2 & 32.4 & 39.7 & 15.3 & 24.4 & 44.2 & 31.7 & 38.9 \\
    \bottomrule
  \end{tabular}
\end{table*}

\begin{table}[t]
  \centering
  \small
  \setlength{\tabcolsep}{3.5pt}
  \caption{\textbf{Video generation models on editing.}}
  \label{tab:s_video}
  \begin{tabular}{@{}lccccccc@{}}
    \toprule
    Model & $\Sedit$ & $C$ & $K$ & $Q$ & $C_\textsc{Wh}$ & $C_\textsc{Wn}$ & $C_\textsc{Hw}$ \\
    \midrule
    MiniMax-H3 & 39.5 & 38.5 & 69.7 & 54.2 & 27.0 & 32.0 & 56.4 \\
    Wan2.1-VACE & 15.6 & 17.2 & 48.0 & 27.1 & 11.9 & 17.0 & 22.0 \\
    \bottomrule
  \end{tabular}
\end{table}

\begin{table}[t]
  \centering
  \small
  \setlength{\tabcolsep}{1.4pt}
  \caption{\textbf{Objective metrics for reconstruction}, with ranks.}
  \label{tab:s_obj}
  \begin{tabular}{@{}lcccc@{}}
    \toprule
    Model & SSIM & DINOv2 & CLIP & $\Srecon$ \\
    \midrule
GPT-6 Astra & \textbf{79.0}\rk{1} & \textbf{80.1}\rk{1} & \textbf{88.1}\rk{1} & \textbf{70.7}\rk{1} \\
GPT-6.1 Sol & \underline{78.3}\rk{2} & \underline{80.0}\rk{2} & 87.3\rk{3} & \underline{69.7}\rk{2} \\
Claude Opus 5.5 & 75.8\rk{4} & 78.5\rk{3} & \underline{87.9}\rk{2} & 63.0\rk{3} \\
GPT-6.1 Terra & 77.0\rk{3} & 77.7\rk{4} & 86.6\rk{5} & 59.7\rk{4} \\
Gemini 4 Argon & 71.7\rk{7} & 73.5\rk{6} & 81.8\rk{8} & 52.3\rk{5} \\
Kimi-K3.1-Pro & 69.9\rk{10} & 70.4\rk{10} & 79.6\rk{12} & 47.1\rk{6} \\
GPT-5.6 Sol & 71.3\rk{8} & 72.1\rk{7} & 83.9\rk{7} & 45.5\rk{7} \\
Claude Haiku 5.5 & 75.2\rk{5} & 74.5\rk{5} & 86.8\rk{4} & 42.2\rk{8} \\
Gemini 3.8 Flash & 67.6\rk{11} & 70.2\rk{11} & 81.4\rk{10} & 40.1\rk{9} \\
GPT-6 Luna & 70.3\rk{9} & 71.0\rk{9} & 81.8\rk{8} & 39.8\rk{10} \\
DeepSeek-V4.1-Flash (1017) & 63.2\rk{12} & 66.9\rk{12} & 79.8\rk{11} & 36.0\rk{11} \\
MiMo-V2.6-Flash & 60.1\rk{15} & 63.5\rk{13} & 74.2\rk{14} & 35.3\rk{12} \\
HY4-preview & 49.1\rk{16} & 48.9\rk{16} & 56.2\rk{16} & 32.7\rk{13} \\
Claude Sonnet 5.5 & 74.5\rk{6} & 72.0\rk{8} & 85.0\rk{6} & 32.7\rk{14} \\
Qwen3.8-27B & 61.7\rk{14} & 61.0\rk{15} & 73.4\rk{15} & 25.9\rk{15} \\
GLM-5.3-Flash & 62.4\rk{13} & 62.3\rk{14} & 74.8\rk{13} & 24.4\rk{16} \\
    \bottomrule
  \end{tabular}
\end{table}

\section{Judge Agreement}
\label{sec:s_rel}

\begin{figure*}[tp]
  \centering
  \includegraphics[width=0.7\textwidth]{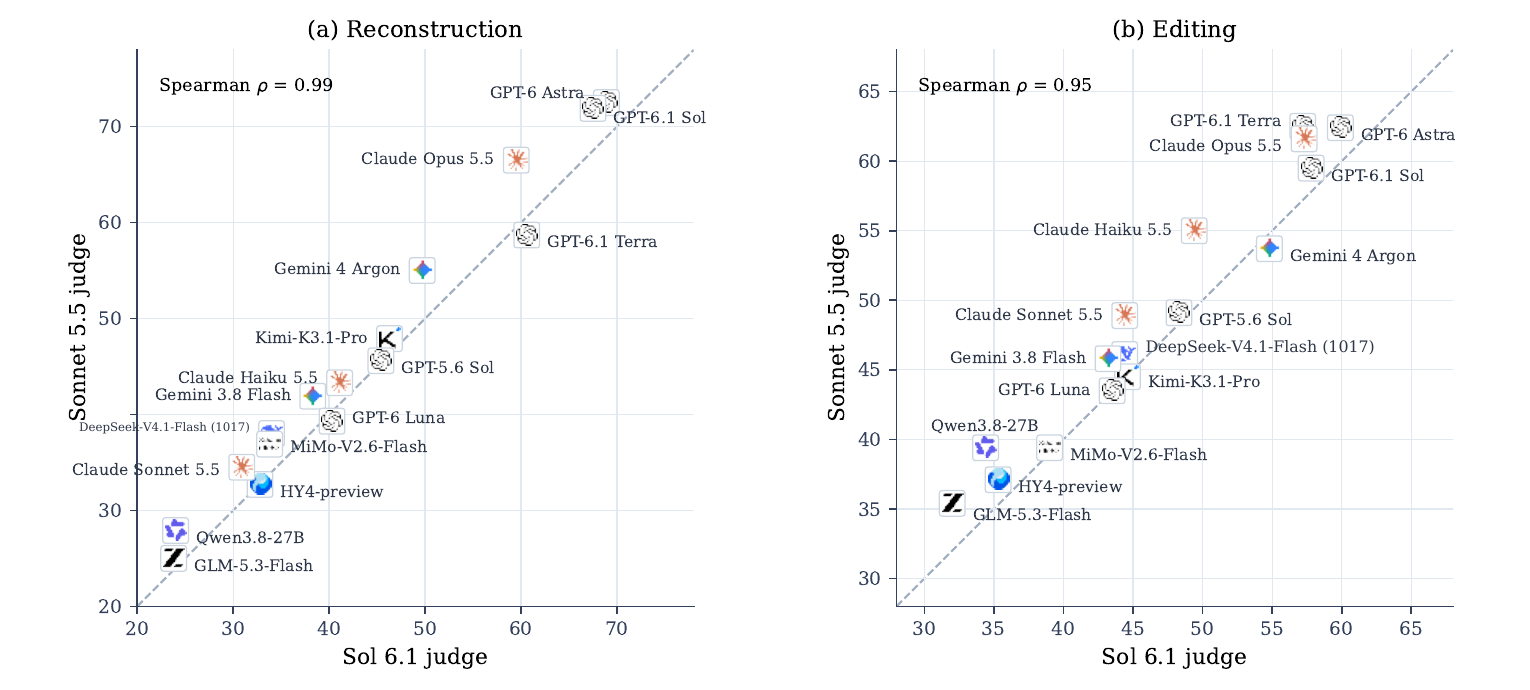}
  \caption{\textbf{Sol 6.1 judge against Sonnet 5.5 judge}, per model.}
  \label{fig:s_judges}
\end{figure*}

\noindent Scores from each judge separately are given in \cref{tab:s_perjudge,tab:s_perjudge_edit,fig:s_judges}.
The two judges rank models almost identically, with Spearman 0.99 for reconstruction and 0.95 for editing, and per clip their reconstruction scores correlate at Pearson 0.61 to 0.81, depending on the model.

\begin{table}[t]
  \centering
  \small
  \setlength{\tabcolsep}{4pt}
  \centering
  \caption{\textbf{Reconstruction scores per judge}, with ranks.}
  \label{tab:s_perjudge}
  \begin{tabular}{@{}lccc@{}}
    \toprule
    Model & Avg. & Sol & Sonnet \\
    \midrule
    GPT-6 Astra & 70.7\rk{1} & 68.9\rk{1} & 72.5\rk{1} \\
    GPT-6.1 Sol & 69.7\rk{2} & 67.5\rk{2} & 71.9\rk{2} \\
    Claude Opus 5.5 & 63.0\rk{3} & 59.5\rk{4} & 66.5\rk{3} \\
    GPT-6.1 Terra & 59.7\rk{4} & 60.6\rk{3} & 58.7\rk{4} \\
    Gemini 4 Argon & 52.3\rk{5} & 49.7\rk{5} & 55.0\rk{5} \\
    Kimi-K3.1-Pro & 47.1\rk{6} & 46.3\rk{6} & 47.9\rk{6} \\
    GPT-5.6 Sol & 45.5\rk{7} & 45.4\rk{7} & 45.6\rk{7} \\
    Claude Haiku 5.5 & 42.2\rk{8} & 41.1\rk{8} & 43.3\rk{8} \\
    Gemini 3.8 Flash & 40.1\rk{9} & 38.3\rk{10} & 41.9\rk{9} \\
    GPT-6 Luna & 39.8\rk{10} & 40.3\rk{9} & 39.3\rk{10} \\
    DeepSeek-V4.1-Flash (1017) & 36.0\rk{11} & 34.0\rk{11} & 38.0\rk{11} \\
    MiMo-V2.6-Flash & 35.3\rk{12} & 33.8\rk{12} & 36.9\rk{12} \\
    HY4-preview & 32.7\rk{13} & 32.8\rk{13} & 32.7\rk{14} \\
    Claude Sonnet 5.5 & 32.7\rk{14} & 30.9\rk{14} & 34.5\rk{13} \\
    Qwen3.8-27B & 25.9\rk{15} & 24.0\rk{15} & 27.9\rk{15} \\
    GLM-5.3-Flash & 24.4\rk{16} & 23.8\rk{16} & 25.0\rk{16} \\
    \bottomrule
  \end{tabular}
\end{table}

\begin{table}[t]
  \centering
  \small
  \setlength{\tabcolsep}{4pt}
  \centering
  \caption{\textbf{Editing scores per judge}, with ranks.}
  \label{tab:s_perjudge_edit}
  \begin{tabular}{@{}lccc@{}}
    \toprule
    Model & Avg. & Sol & Sonnet \\
    \midrule
    GPT-6 Astra & 61.2\rk{1} & 59.9\rk{1} & 62.4\rk{2} \\
    GPT-6.1 Terra & 59.9\rk{2} & 57.2\rk{4} & 62.5\rk{1} \\
    Claude Opus 5.5 & 59.5\rk{3} & 57.3\rk{3} & 61.6\rk{3} \\
    GPT-6.1 Sol & 58.6\rk{4} & 57.8\rk{2} & 59.5\rk{4} \\
    Gemini 4 Argon & 54.2\rk{5} & 54.8\rk{5} & 53.7\rk{6} \\
    Claude Haiku 5.5 & 52.2\rk{6} & 49.4\rk{6} & 55.0\rk{5} \\
    GPT-5.6 Sol & 48.7\rk{7} & 48.3\rk{7} & 49.1\rk{7} \\
    Claude Sonnet 5.5 & 46.7\rk{8} & 44.4\rk{9} & 48.9\rk{8} \\
    DeepSeek-V4.1-Flash (1017) & 45.2\rk{9} & 44.4\rk{10} & 46.1\rk{9} \\
    Kimi-K3.1-Pro & 44.6\rk{10} & 44.6\rk{8} & 44.5\rk{11} \\
    Gemini 3.8 Flash & 44.5\rk{11} & 43.2\rk{12} & 45.8\rk{10} \\
    GPT-6 Luna & 43.5\rk{12} & 43.5\rk{11} & 43.5\rk{12} \\
    MiMo-V2.6-Flash & 39.2\rk{13} & 39.0\rk{13} & 39.4\rk{13} \\
    Qwen3.8-27B & 36.9\rk{14} & 34.4\rk{15} & 39.4\rk{14} \\
    HY4-preview & 36.2\rk{15} & 35.3\rk{14} & 37.1\rk{15} \\
    GLM-5.3-Flash & 33.7\rk{16} & 32.0\rk{16} & 35.4\rk{16} \\
    \bottomrule
  \end{tabular}
\end{table}

\FloatBarrier
\clearpage
\onecolumn
\section{Prompts}
\label{sec:s_prompts}

\noindent Variables in \phl{red braces} are filled in per clip or per task.
Box colors: \textcolor{rcol}{\textbf{red}} for the models under test and \textcolor{gcol}{\textbf{green}} for the judges.

\begin{figure}[!htbp]
\begin{tcblisting}{promptstyle=rcol, title={Reconstruction task prompt}}
You are given a short reference video as (*@\phl{n\_frames}@*) frames (timestamps in the list below).
Video facts: duration (*@\phl{duration}@*) s, size (*@\phl{width}@*)x(*@\phl{height}@*).
Frames: (*@\phl{frame\_list}@*)

Task: write code that reproduces this video as faithfully as possible: the same elements, appearance, layout, text, motion paths, event order and timing, and final state. You never see any original code; infer everything from the frames.

Output rules (the video is recorded automatically):
- You may use HTML, CSS, SVG, Canvas 2D, WebGL, or three.js, with no network access.
- No external images, fonts or media. Approximate photos or footage with shapes, gradients and text.
- Animation must be a deterministic function of time (no Math.random without a fixed seed, no real-time clock).

Return only the complete index.html inside one ```html code block.
\end{tcblisting}
\caption{\textbf{Reconstruction task prompt.} Given to each code model with the reference frames; it returns one self-contained \texttt{index.html}.}
\label{fig:p_recon}
\end{figure}

\begin{figure}[!htbp]
\begin{tcblisting}{promptstyle=gcol, title={Reconstruction judge}}
You judge how well a CANDIDATE video reproduces a REFERENCE video. Both are given as frames at the same timestamps.
Reference frames: (*@\phl{ref\_frames}@*)
Candidate frames: (*@\phl{cand\_frames}@*)
Rubric (questions are about the CANDIDATE, decided by comparing with the reference): (*@\phl{rubric\_json}@*)
1. For every rubric item, answer pass (1) or fail (0) with a one-line reason.
   - Be strict: partial = fail.
   - Allow small timing offsets (+/-0.5 s) and small color or size differences that a viewer would not notice.
2. Give holistic scores from 1 to 5:
   - appearance: elements, colors, layout, text;
   - motion: paths, timing, order;
   - overall: would a viewer say it is the same video?;
   - quality: artifacts, clipping, flicker or broken rendering in the candidate itself.
Return only JSON: {"items": [{"id": "r1", "pass": 1, "reason": "..."}], "holistic": {"appearance": 4, "motion": 3, "overall": 3, "quality": 5}, "summary": "one sentence"}
\end{tcblisting}
\caption{\textbf{Reconstruction judge.} Sol 6.1 and Sonnet 5.5 see reference and candidate frames at the same times plus the frozen rubric; the overall rating is the Fidelity score of \cref{tab:recon}.}
\label{fig:p_rjudge}
\end{figure}

\begin{figure}[!htbp]
\begin{tcblisting}{promptstyle=rcol, title={Editing task prompt}}
The current directory contains index.html, a single-file reconstruction (HTML/CSS/JS) of a reference video. Frames of the reference video and of the current reconstruction's rendering are attached, with their times in seconds. The recording rules of the reconstruction task still hold; keep the canvas size and total duration unchanged.
EDIT REQUEST (times refer to the video): (*@\phl{instr}@*)
Edit index.html in place so that its rendering shows this change. Change only what the request asks for and keep everything else as it is. The edit must be clearly visible.
\end{tcblisting}
\caption{\textbf{Editing task prompt.} The model edits its own reconstruction (code and rendered frames), given reference frames and the instruction; it does not see the checks or keep items.}
\label{fig:p_edit}
\end{figure}

\begin{figure}[!htbp]
\begin{tcblisting}{promptstyle=gcol, title={Edit judge}}
You judge a video EDIT. A model was asked to edit a video according to this request:
"(*@\phl{instr}@*)"
You see frames of the ORIGINAL video (REF) and of the edited result (AFTER) at the same times. For each change, frames inside its time window are listed first. Images: (*@\phl{listing}@*)
Changes and check questions (decide each question from the AFTER frames; use REF to understand what the instruction refers to and what the original looks like):
(*@\phl{changes}@*)
Must stay unchanged (decide if each is still the same in AFTER as in REF): (*@\phl{must}@*)
Rules: if nothing relevant is visibly different from the original in a change's window, mark every check of that change "missed". "partial" = the change is there but wrong in amount, place or timing. Be strict and concrete. Also give quality 1-5 for AFTER only (5 = clean; 1 = broken, flicker, clipping, overlap, blank).
Return ONLY JSON: {"changes": [{"i": 0, "visible_change": true, "checks": [{"verdict": "done|partial|missed", "reason": "..."}]}], "must_stay": [{"item": "...", "kept": true, "reason": "..."}], "quality": 4, "summary": "one sentence"}
\end{tcblisting}
\caption{\textbf{Edit judge.} Sol 6.1 and Sonnet 5.5 see reference and edited frames at the same times.}
\label{fig:p_ejudge}
\end{figure}

\clearpage
\section{Qualitative Cases}
\label{sec:s_cases}

\begin{figure}[!h]
  \centering
  \includegraphics[width=\textwidth,height=0.86\textheight,keepaspectratio]{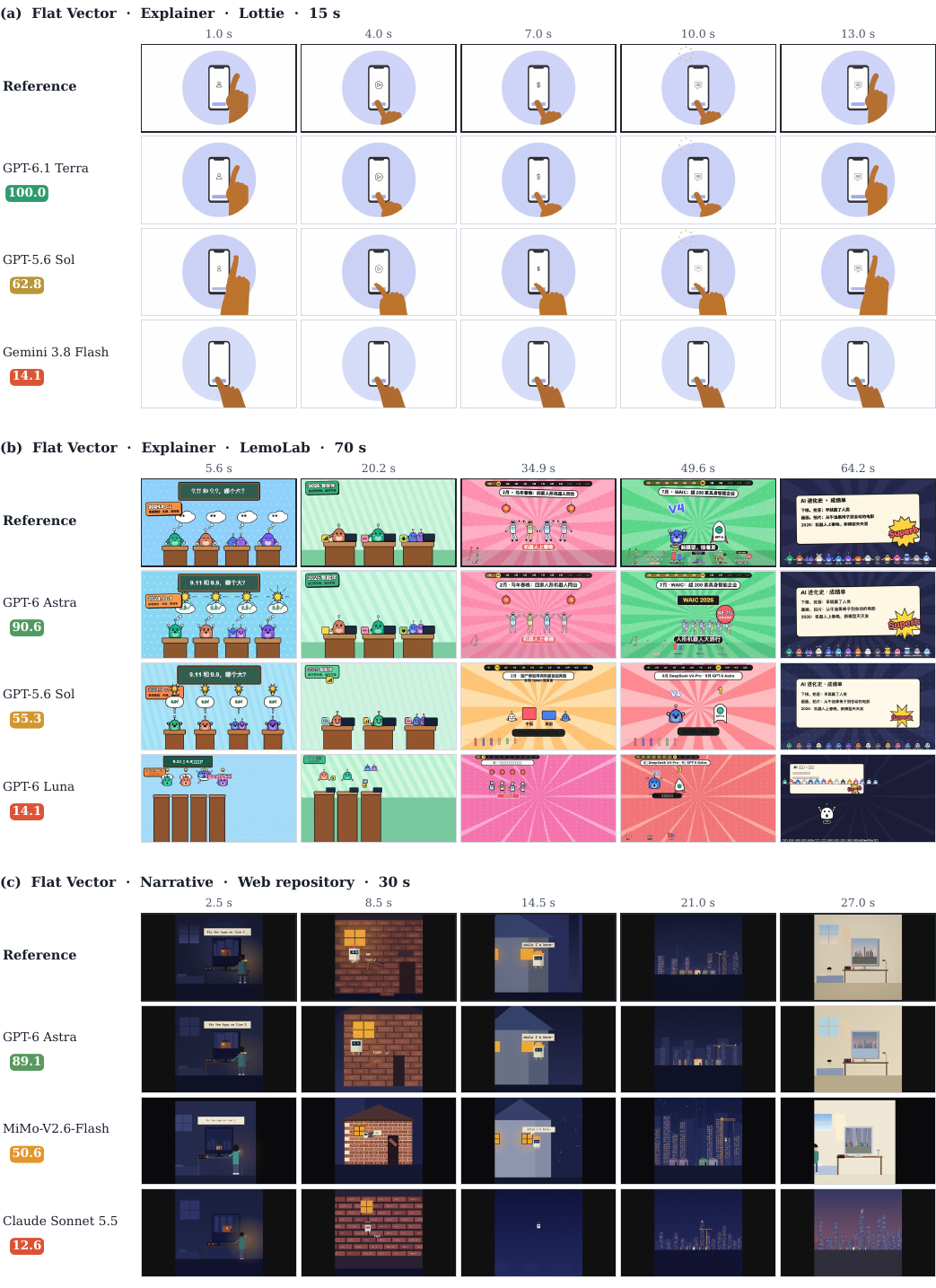}
  \caption{\textbf{Reconstruction cases: Flat Vector.} Each block: reference and three reconstructions at the same five times; chips give $S_\mathrm{recon}$.}
  \label{fig:case_r1}
\end{figure}

\begin{figure*}[p]
  \centering
  \includegraphics[width=\textwidth,height=0.93\textheight,keepaspectratio]{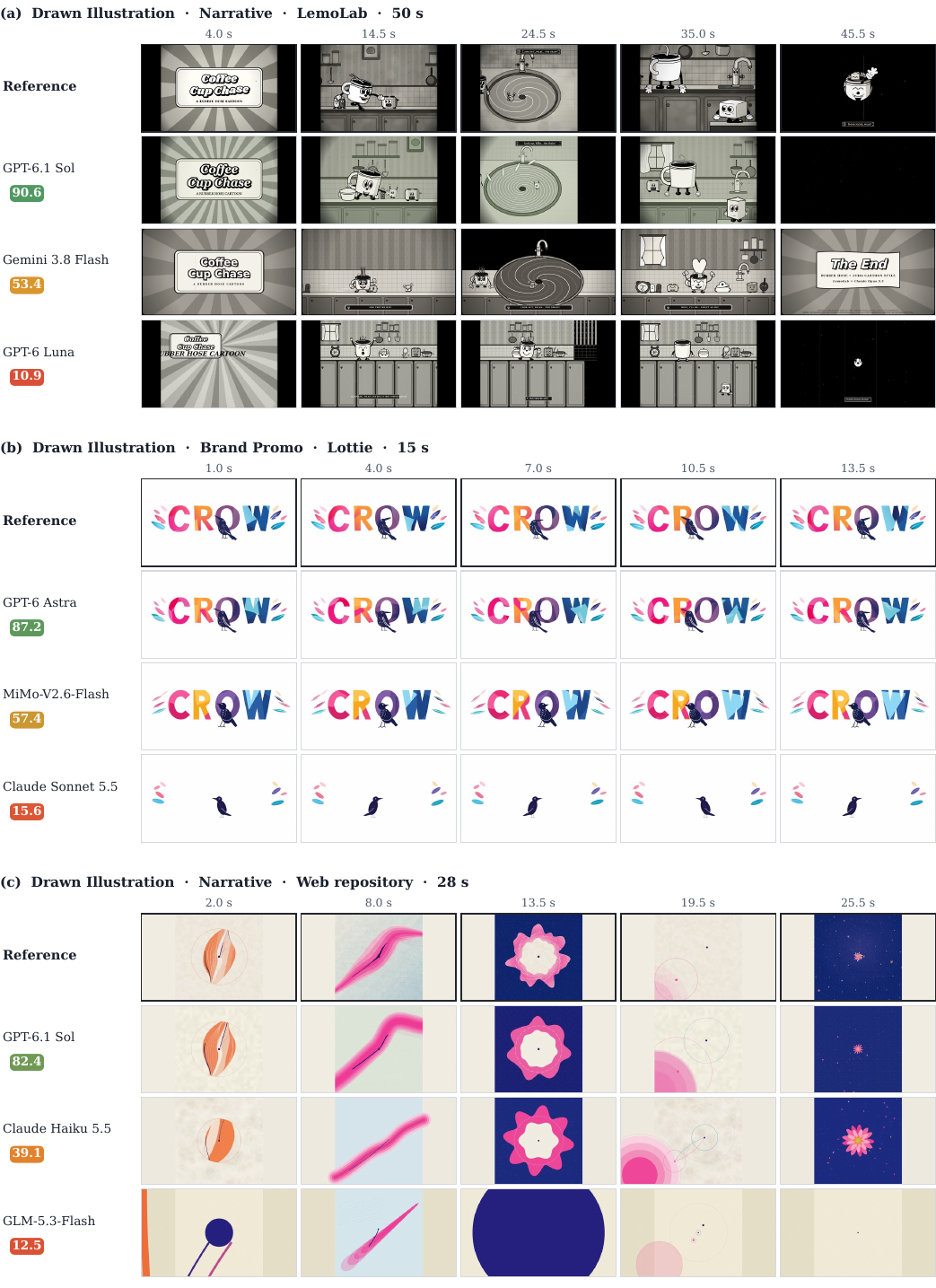}
  \caption{\textbf{Reconstruction cases: Drawn Illustration.}}
  \label{fig:case_r2}
\end{figure*}

\begin{figure*}[p]
  \centering
  \includegraphics[width=\textwidth,height=0.93\textheight,keepaspectratio]{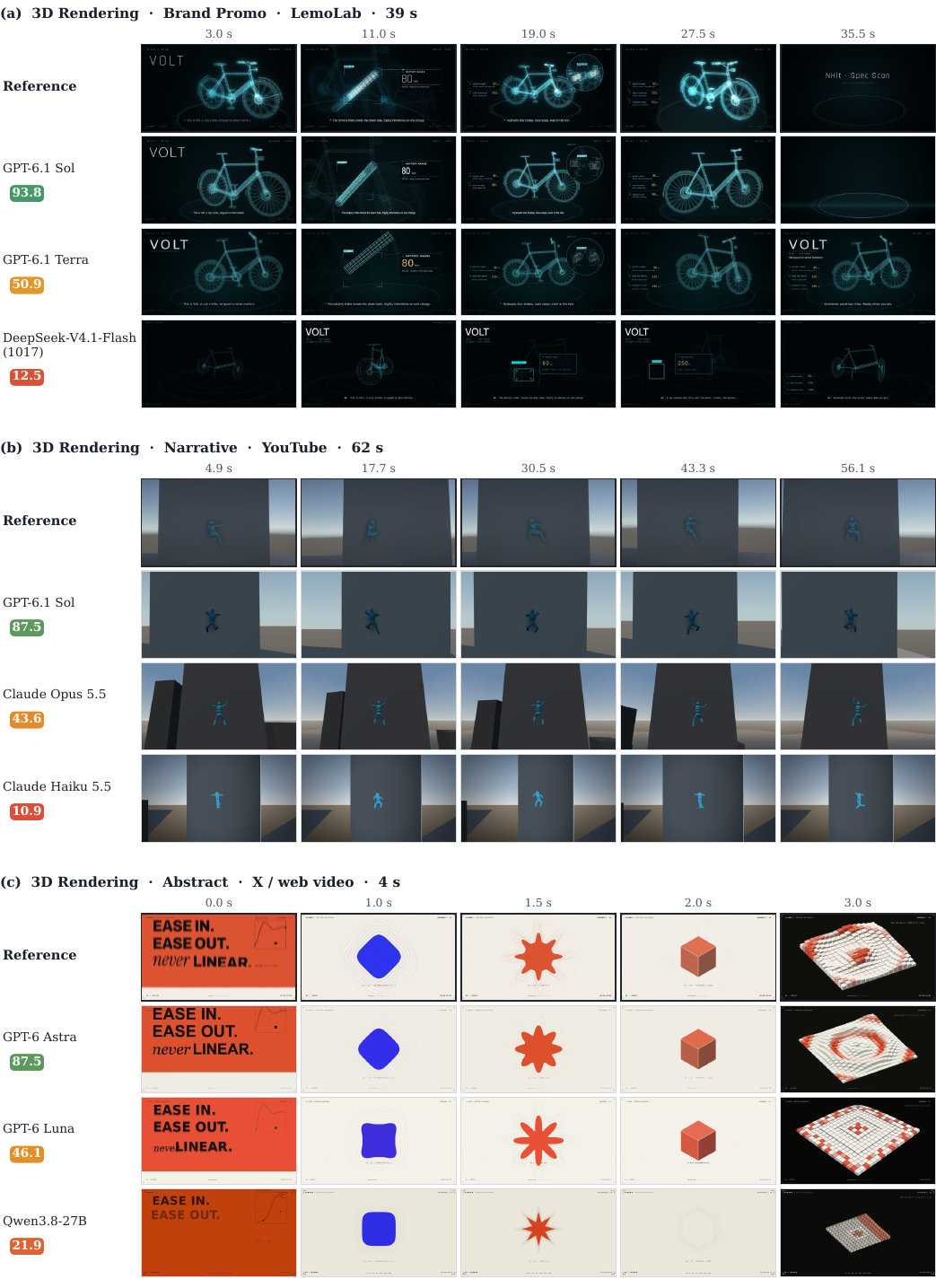}
  \caption{\textbf{Reconstruction cases: 3D Rendering.}}
  \label{fig:case_r3}
\end{figure*}

\begin{figure*}[p]
  \centering
  \includegraphics[width=\textwidth,height=0.93\textheight,keepaspectratio]{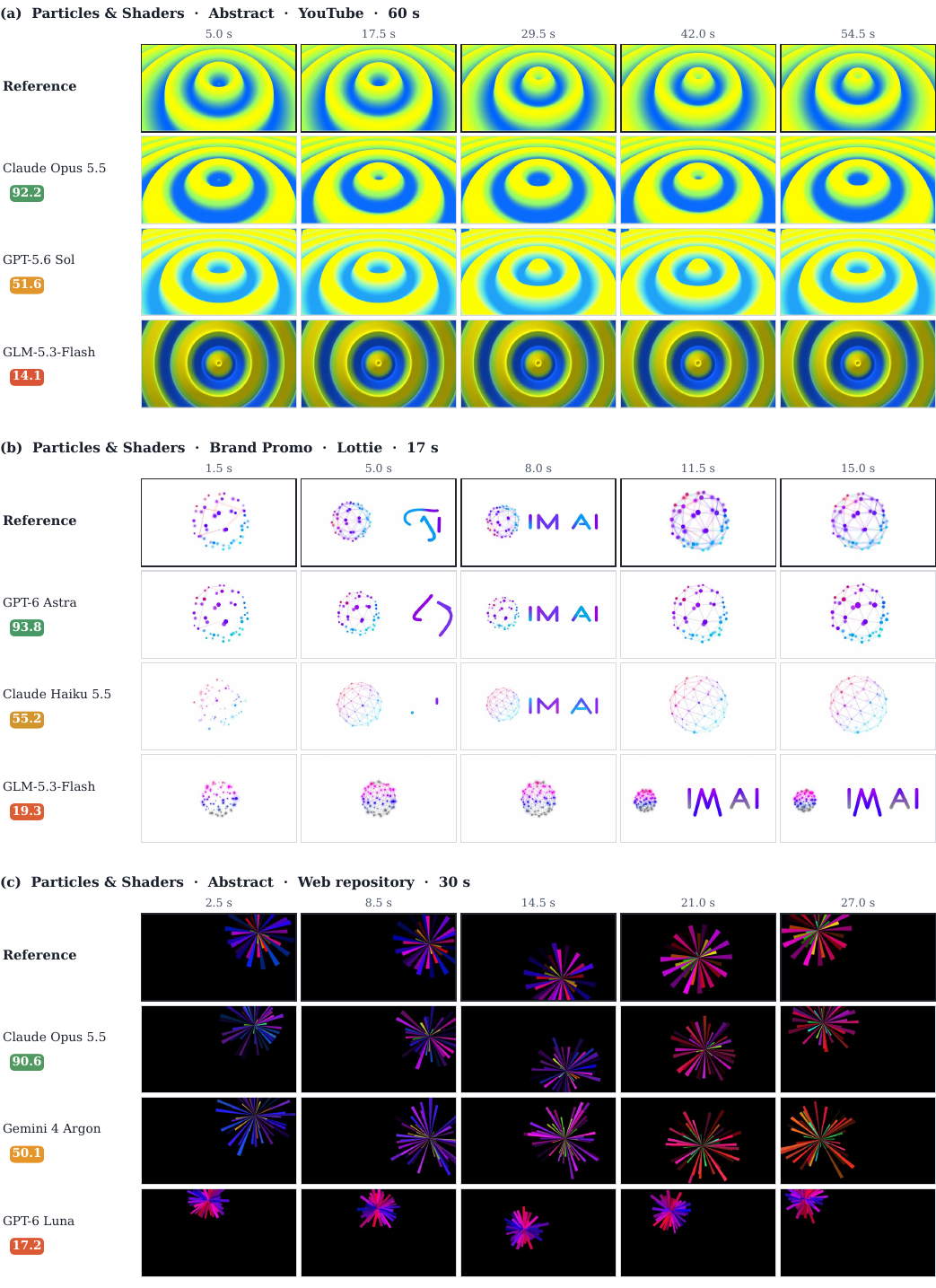}
  \caption{\textbf{Reconstruction cases: Particles \& Shaders.}}
  \label{fig:case_r4}
\end{figure*}

\begin{figure*}[p]
  \centering
  \includegraphics[width=\textwidth,height=0.93\textheight,keepaspectratio]{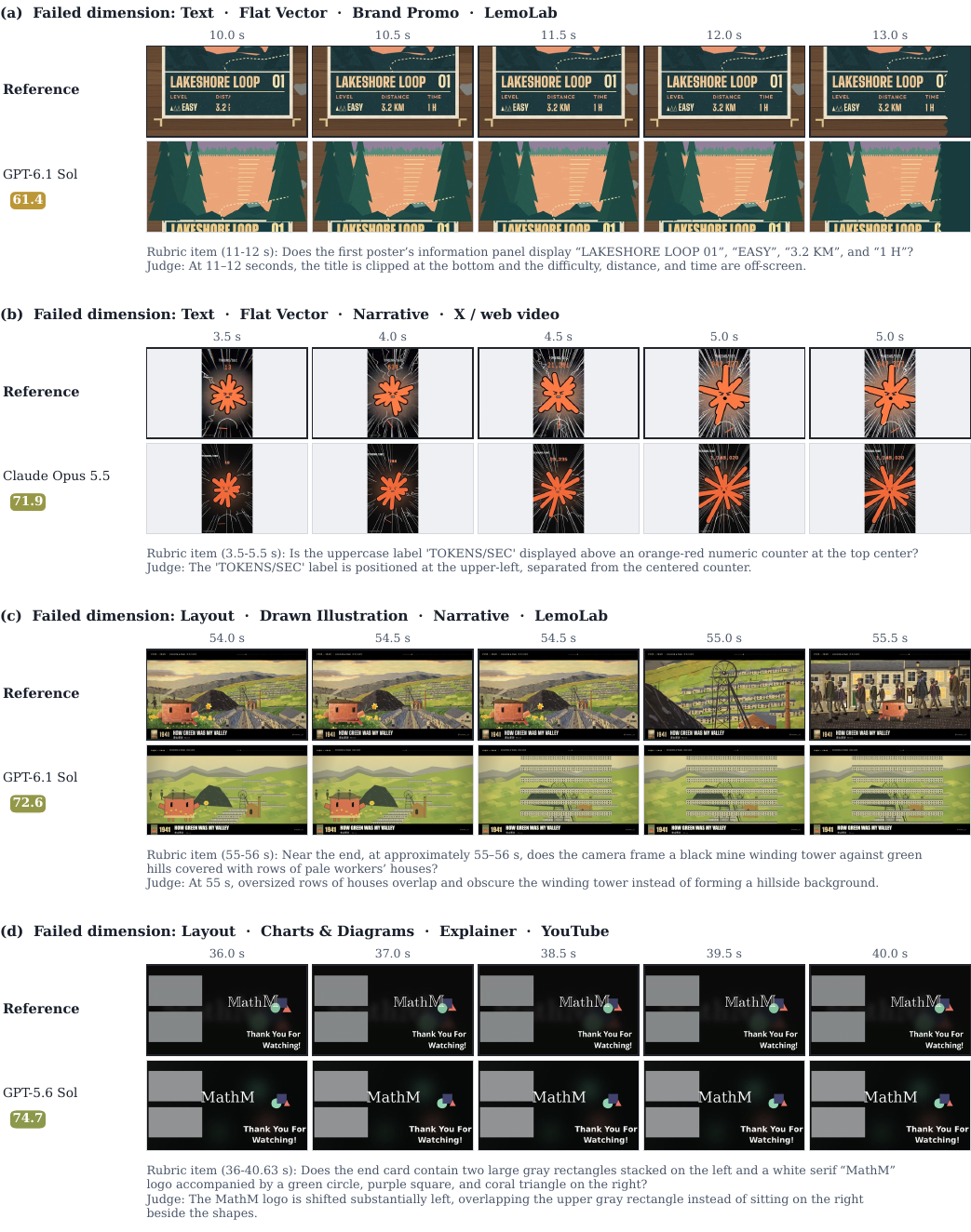}
  \caption{\textbf{Reconstruction failures: Text and layout.} Each reconstruction fails only the named rubric dimension. The failed rubric item and the judge's reason are quoted under each pair.}
  \label{fig:case_d1}
\end{figure*}

\begin{figure*}[p]
  \centering
  \includegraphics[width=\textwidth,height=0.93\textheight,keepaspectratio]{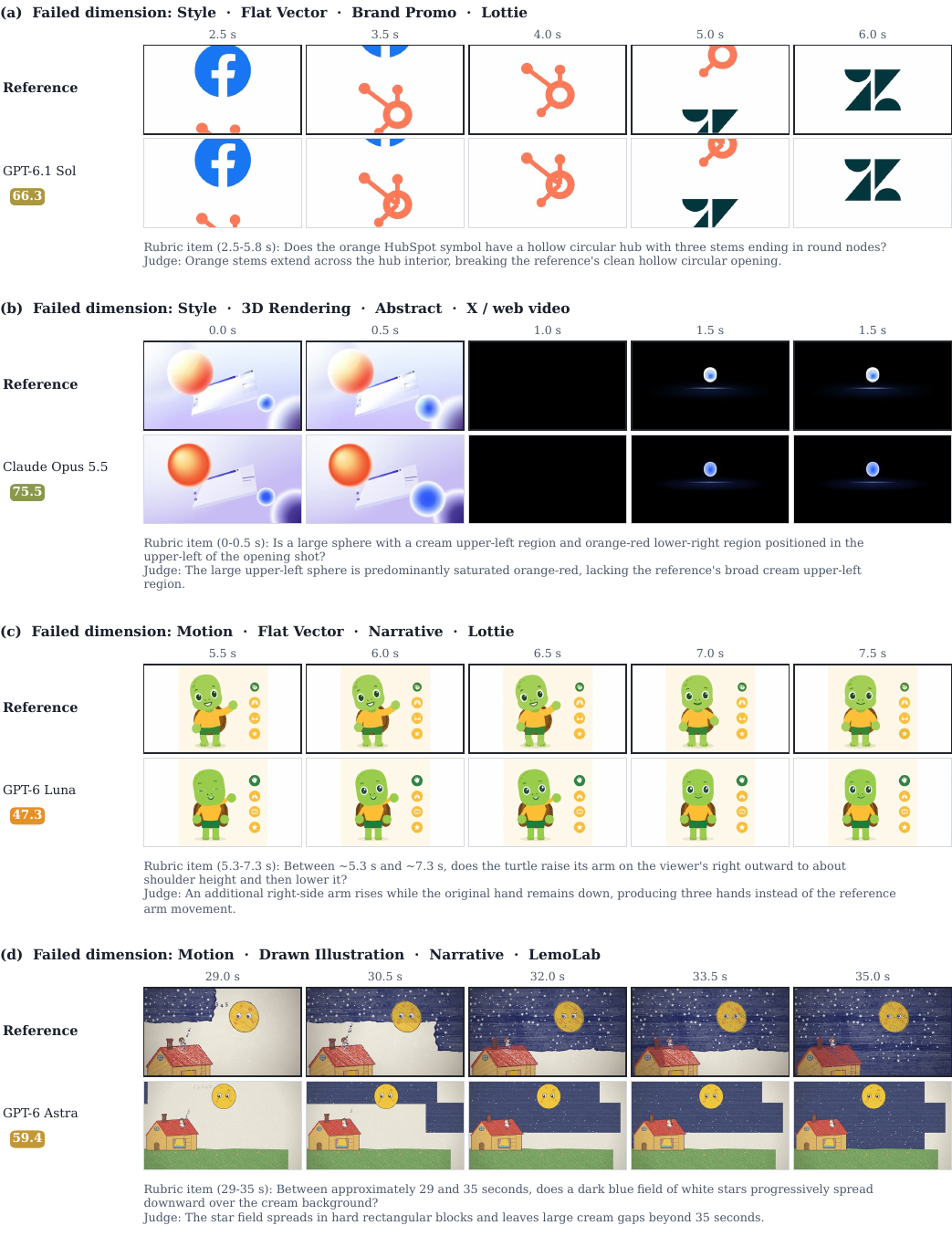}
  \caption{\textbf{Reconstruction failures: Style and motion.}}
  \label{fig:case_d2}
\end{figure*}

\begin{figure*}[p]
  \centering
  \includegraphics[width=\textwidth,height=0.93\textheight,keepaspectratio]{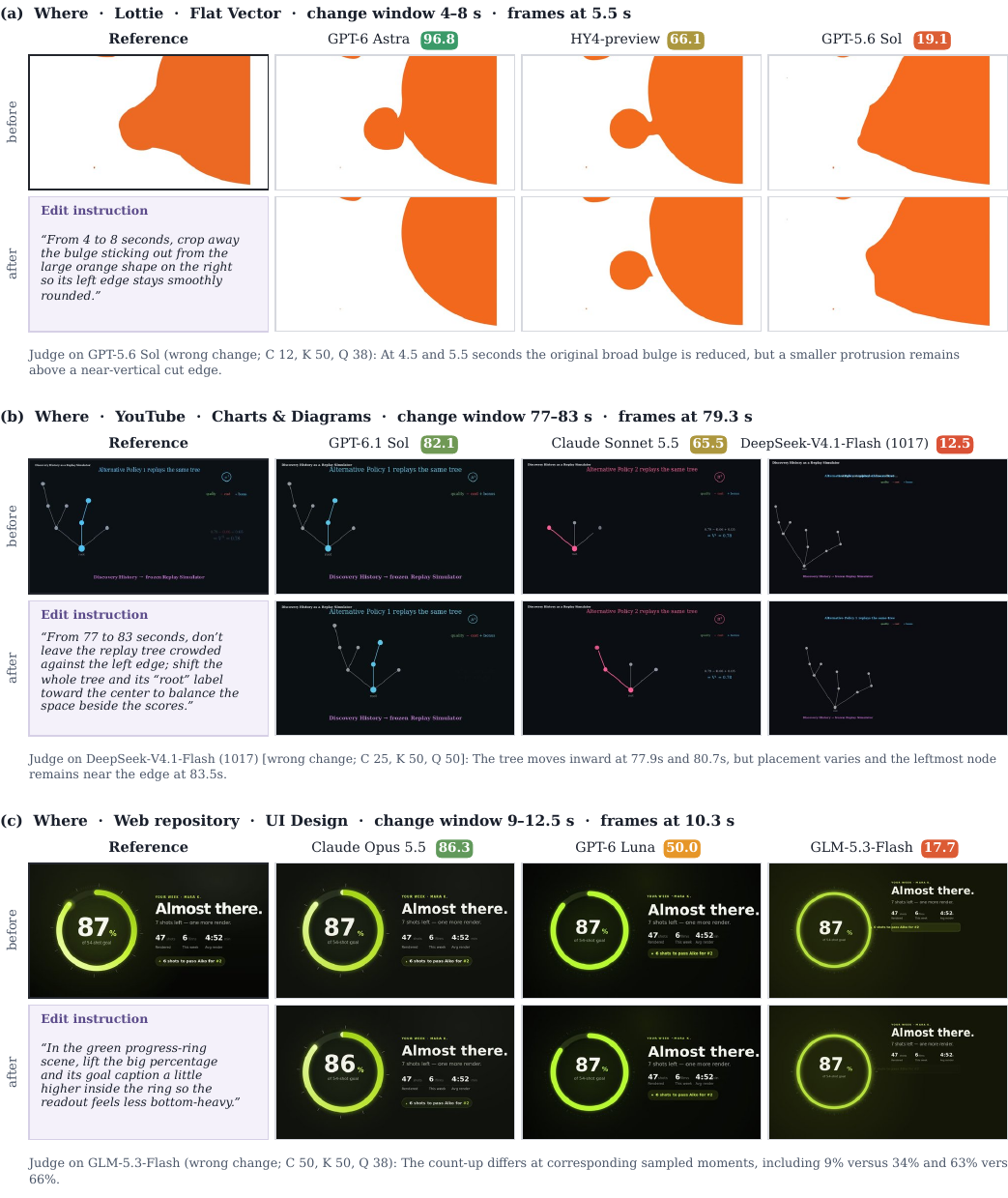}
  \caption{\textbf{Editing cases: \where changes.} (a)--(c) \where. Reference column: original frame and edit instruction; other columns: each model's reconstruction (top) and edit (bottom) at the same time. Chips give $S_\mathrm{edit}$; notes quote the judge on the weakest model.}
  \label{fig:case_e1}
\end{figure*}

\begin{figure*}[p]
  \centering
  \includegraphics[width=\textwidth,height=0.93\textheight,keepaspectratio]{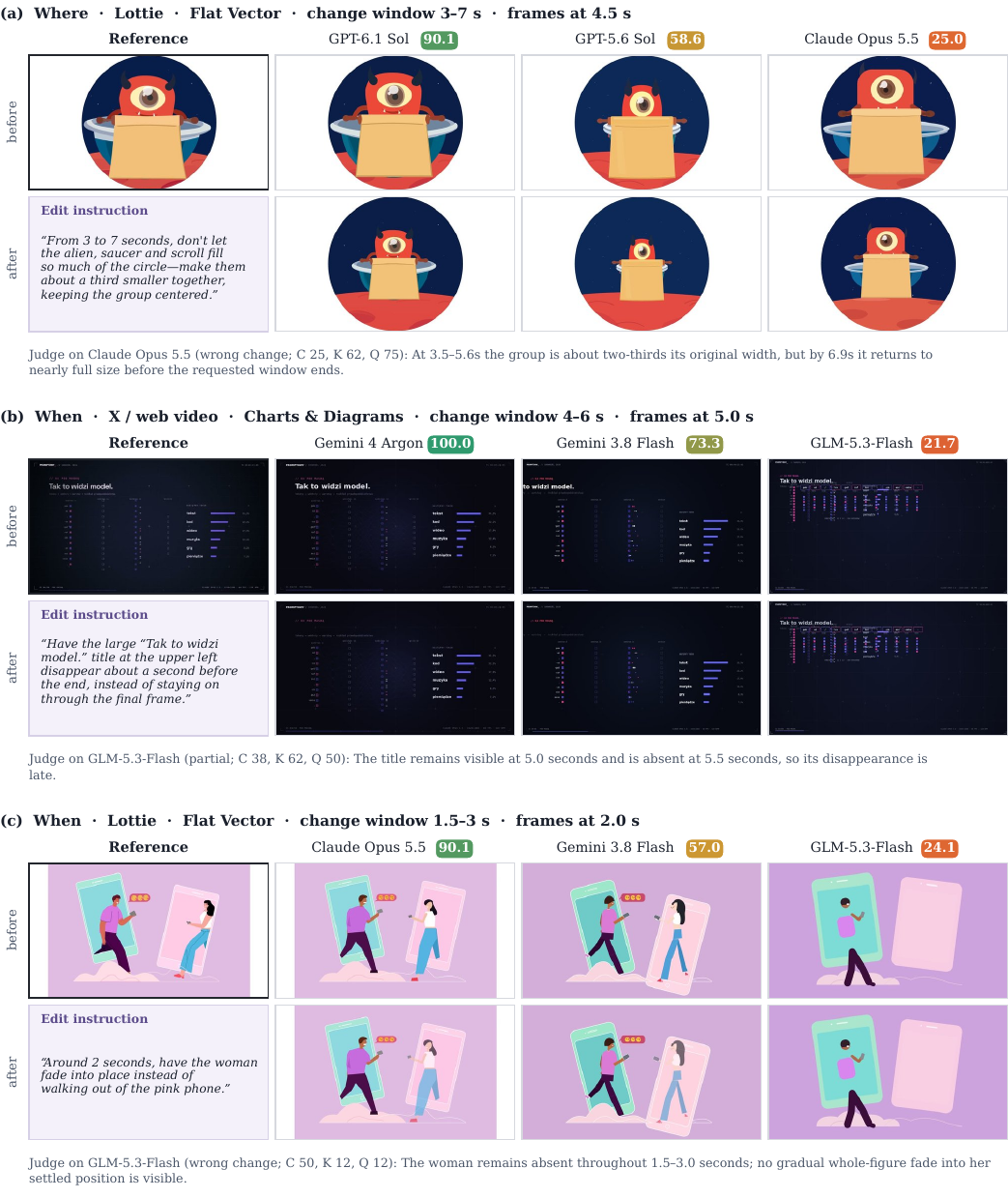}
  \caption{\textbf{Editing cases: \where and \when changes.} (a) \where; (b), (c) \when.}
  \label{fig:case_e2}
\end{figure*}

\begin{figure*}[p]
  \centering
  \includegraphics[width=\textwidth,height=0.93\textheight,keepaspectratio]{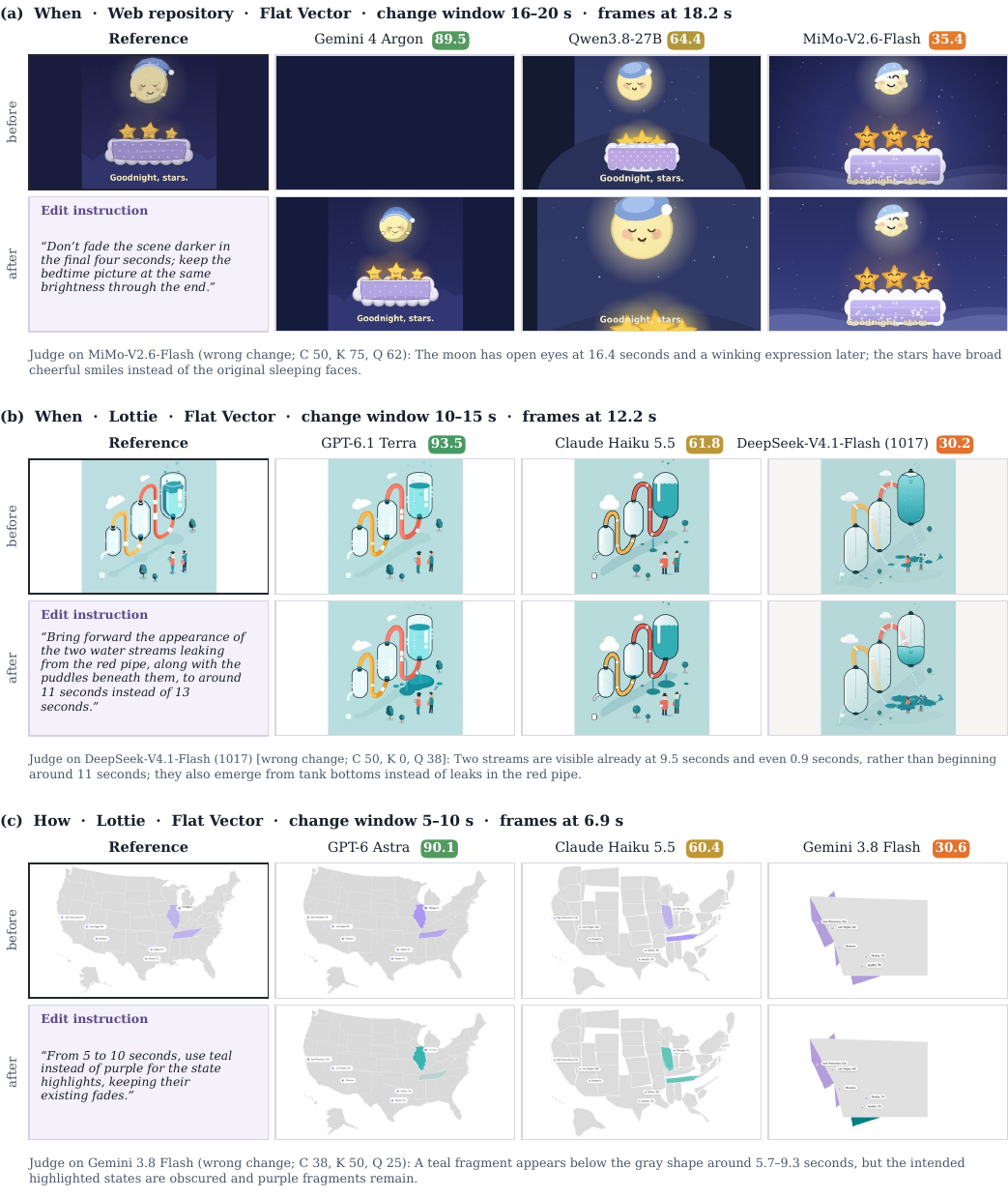}
  \caption{\textbf{Editing cases: \when and \how changes.} (a), (b) \when; (c) \how.}
  \label{fig:case_e3}
\end{figure*}

\begin{figure*}[p]
  \centering
  \includegraphics[width=\textwidth,height=0.93\textheight,keepaspectratio]{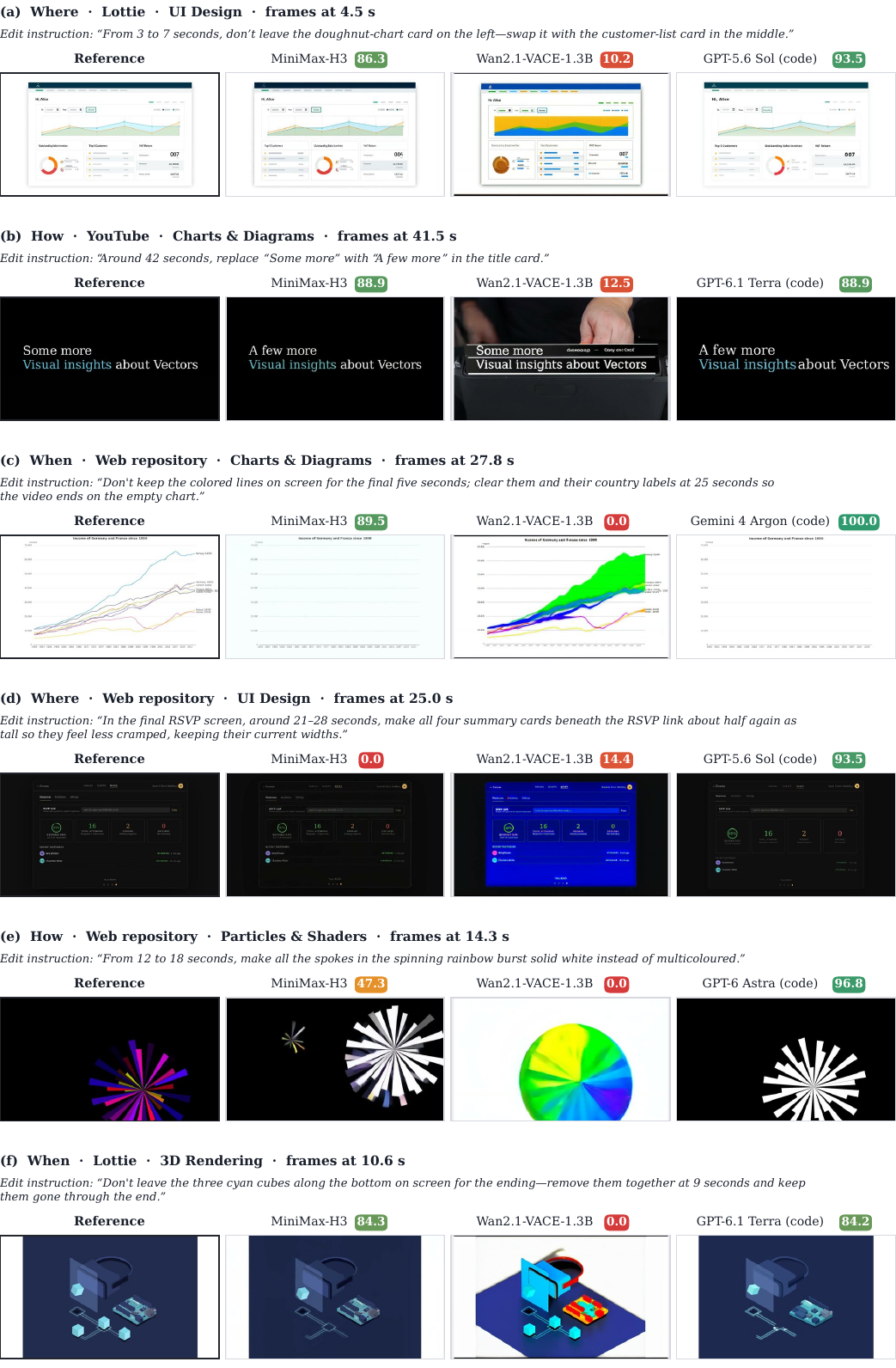}
  \caption{\textbf{Video generation baselines.} The edited frame from MiniMax-H3, Wan2.1-VACE-1.3B, and the best code model on the same task, with $S_\mathrm{edit}$ for each.}
  \label{fig:case_v1}
\end{figure*}

\looseness=-1 \noindent Frame strips come from the runs scored in the main tables, and score chips give the two-judge mean (green high, red low).
\Cref{fig:case_r1,fig:case_r2,fig:case_r3,fig:case_r4} show reconstructions across visual forms, \cref{fig:case_d1,fig:case_d2} reconstructions that fail a single rubric dimension, \cref{fig:case_e1,fig:case_e2,fig:case_e3} edits by change type, and \cref{fig:case_v1} the video generation baselines.

\end{document}